\documentclass[journal]{IEEEtran}
\pdfoutput=1
\ifCLASSINFOpdf
\else
\fi

\usepackage{amsmath}
\usepackage{algpseudocode}

\usepackage{hyperref}
\usepackage{graphicx}
\usepackage{booktabs}

\usepackage{adjustbox}
\usepackage{subfig}
\usepackage{wrapfig}

\usepackage[font={footnotesize}]{caption}
\usepackage[ruled,vlined]{algorithm2e}

\SetKwInput{kwTerms}{Terms}

\SetKwInput{kwTrain}{Train the Encoder / Decoder}

\SetKwInput{kwGen}{Generate Samples}
\SetKwInput{kwForm}{Formulas}

\usepackage{tikz}

\usetikzlibrary{arrows.meta,
                chains,
                positioning,
                shapes.geometric
                }

\usepackage{lipsum,multicol}

\begin{document}
%
\title{Towards Understanding How Data Augmentation Works with Imbalanced Data}

\author{Damien~A.~Dablain
        and~Nitesh~V.~Chawla $^{\dagger}$,~\IEEEmembership{IEEE, Fellow}
\thanks{$^{\dagger}$ corresponding author}
\thanks{D. Dablain and N.V. Chawla are with the Dept. Computer Science and Engineering \& Lucy Family Institute for Data and Society at the University of Notre Dame, Notre Dame, IN 46556 e-mail: \{ddablain,nchawla\}@nd.edu}
}

\maketitle

\begin{abstract}
Data augmentation forms the cornerstone of many modern machine learning training pipelines; yet, the mechanisms by which it works are not clearly understood. Much of the research on data augmentation (DA) has focused on improving existing techniques, examining its regularization effects in the context of neural network over-fitting, or investigating its impact on features. Here, we undertake a holistic examination of the effect of DA on three different classifiers, convolutional neural networks, support vector machines, and logistic regression models, which are commonly used in supervised classification of imbalanced data. We support our examination with testing on three image and five tabular datasets. Our research indicates that DA, when applied to imbalanced data, produces substantial changes in model weights, support vectors and feature selection; even though it may only yield relatively modest changes to global metrics, such as balanced accuracy or F1 measure. We hypothesize that DA works by facilitating variances in data, so that machine learning models can associate changes in the data with labels. By diversifying the range of feature amplitudes that a model must recognize to predict a label, DA improves a model's capacity to generalize when learning with imbalanced data. 
\end{abstract}


\IEEEpeerreviewmaketitle

\section{Introduction}
\label{sec:int}

\IEEEPARstart{D}{ata} Augmentation (DA) \cite{baird1992document,yaeger1996effective} is widely used throughout many machine learning (ML) training regimes, including: computer vision \cite{shorten2019survey}, speech recognition \cite{ko2015audio}, natural language processing \cite{li2022data}, adversarial example training \cite{madry2017towards}, contrastive learning \cite{chen2020simple,he2020momentum}, imbalanced learning \cite{chawla2002smote}, and the microbiome \cite{gordon2022data}. The ubiquity of DA is largely due to its beneficial impact on generalization - the ability of ML models to make correct predictions with respect to data unseen during training \cite{shorten2019survey}. 
It has also been demonstrated to convey other benefits, such as improving robustness to transformations in data and model calibration \cite{zhang2017mixup}. In addition, it can expand the quantity of training data without incurring extra time to secure and label natural data.

Much of the research on DA has focused on enhancing the underlying model's generalization properties, as measured by its ability to improve accuracy metrics, such as balanced accuracy (BAC) and macro F1 measure. Leading techniques to \textit{improve} accuracy metrics have been developed in imbalanced learning \cite{chawla2002smote,he2008adasyn,bellinger2020remix} and balanced image classification \cite{zhang2017mixup,hendrycks2019augmix,yun2019cutmix}. Several papers have also described the regularization effects of DA in deep architectures \cite{bishop1995training,hernandez2018data,carratino2020mixup} and its effect on feature selection \cite{shen2022data,ilse2021selecting,allen2022feature}. However, limited research has been conducted on the holistic impact of DA on \textit{both} single layer and multi-layer ML models. 

We conduct our analysis of the effects of DA on ML models with imbalanced data. Imbalanced data provides a convenient venue for examining the impact of DA on ML models because there are clear differences in the number of class training instances. With imbalanced data, the majority class(es) have more training instances than the minority. Therefore, the impact of DA can be clearly observed and isolated because DA is applied more aggressively with respect to minority classes to synthetically increase their number.

Krizhevsky et al. \cite{krizhevsky2017imagenet} famously observed that DA improved the top-1 error rate on ImageNet when training with AlexNet by over 1\%. We are inspired by this observation to ask the general question: given a relatively modest improvement in accuracy, what is the quantum of change that occurred in the ML \textit{model itself} as a result of DA? Here, we undertake a study of how DA works \textit{underneath the surface} of ML models. We examine a range of ML models involved in supervised classification, including Support Vector Machines (SVM), Logistic Regression (LG), and Convolutional Neural Networks (CNN) with eight different datasets. 




To better understand how DA works when learning with imbalanced data, we design several research questions (RQs). Our RQs investigate the impact of class numerical equalization and feature manipulation on model accuracy; and whether the choice of where to augment - in real or latent space - is affected by data type. Our RQs also consider how DA affects model weights, support vectors and front-end feature selection.

\noindent \textbf{Contributions.}  
In the process of better understanding how DA works, we discover the {\bf{following insights}} towards enhancing the understanding of role of DA within the context of imbalanced data:  

\begin{itemize}
    
    \item \textbf{Simple random over-sampling contributes a greater share to BAC and F1 improvement than feature magnitude manipulation with interpolation.} Even though the minority class features may be noisy compared to the training feature distribution, ML models struggle to accurately predict classes with few samples. This is likely due to the nature of empirical risk minimization (ERM) \cite{vapnik1999nature}, which is employed in the training of many modern ML models. ERM seeks to minimize a \textit{global} or \textit{average} measure of error. This approach over-weights the majority. We find that simple numerical balancing of classes, through random over-sampling or basic cost-reweighting, contributes a greater share to BAC and F1 improvement than feature manipulation with interpolation, when learning with imbalanced data. 
    
    
    
    \item \textbf{The decision of where to augment - real or model latent space - may be affected by data type.} We find that DA in latent space is more effective when dealing with image data than tabular data. This may be due to the nature of the data itself. Tabular data often has a lower dimension than image data. It also may already contain a compact encoding. Because the data representation is already compressed, injecting noise, or changes in the \textit{amplitude} of features in latent space, may confuse ML models trained to distinguish feature amplitudes. 
    
    \item \textbf{In LG and CNN models, DA induces large changes in model weights.} We observe substantial changes (in excess of 100\%) in model weights as a result of DA which simply \textit{copies} minority class instances, or selects minority instances and makes \textit{linear} adjustments to feature amplitudes. Both simple instance copying and linear feature adjustments have the effect of changing the distribution of input feature \textit{amplitudes}. In other words, they change the \textit{average feature amplitude} of features of training sets used by ML models trained to minimize \textit{average ERM}.
    
    \item \textbf{In LG and CNN models trained with imbalanced data, DA causes meaningful changes in both majority and minority class \textit{front-end feature selection}}. This applies whether DA is implemented as a pre-processing step or later in model latent space. We track the most salient features that LG, NN and CNN models use when making model decisions and call them top-K features. We liken the adjustment to \textit{front-end} feature selection as a form of top-down attention adjustment, when DA is implemented in latent space,
    
    \item \textbf{In SVM models, DA induces large changes in the number of support vectors.} Unlike LG and NN models, which accept features as input, SVM models store discriminative instances (support vectors). We observe large changes in the number of support vectors, which are analogous to features in LG and NN models, when models are trained with DA. Some of the support vectors added through DA are attributable to synthetic instances. 

    \item \textbf{More broadly, we hypothesize that DA works by changing the distribution of feature amplitudes in both real and latent space. This, in turn, modifies weight activation and feature (or support vector) selection.} Outside the field of imbalanced learning, researchers have observed that a variety of techniques increase generalization in parametric ML models through weight regularization. These include weight decay \cite{hanson1988comparing}, drop-out \cite{srivastava2014dropout}, data augmentation, and Gaussian noise injection. These mechanisms traditionally work at various points in ML model processing: input features (DA), latent features (drop-out by zeroing weights, which zeros features), and weight decay (changes impact of weights on latent features). We hypothesize that they all share a common trait - they vary the amplitudes of input or latent features. Diversifying the amplitude of features associated with a label improves generalization and robustness because it allows a ML model to connect signals with differing amplitudes with labels.
    
\end{itemize}


\section{Background \& related work}
\label{sec:brw}
In this section, we provide both background to our approach and a discussion of work that relates to our main concepts. We start with an overview of the central problems posed by imbalanced data and how the field has addressed them. We then seek to place these problems, and one particular approach to them (over-sampling or DA for the minority class) in the broader context of machine learning.

We then provide an overview of the theories that describe data augmentation's influence on machine learning models. Much of this research has been conducted outside of imbalanced learning. Within the field, much of the research related to how DA works has been performed on a single algorithm, SMOTE \cite{chawla2002smote}. We also discuss DA as real and latent feature (amplitude) manipulation.

\subsection{Imbalanced learning} Imbalanced learning focuses on how a disparity in the number of class samples affects the training of supervised classifiers. The classes are  colloquially referred to as the majority class(es) (with more samples) and the minority class(es) (with fewer samples). The number of majority and minority classes varies based on whether the labels are binary or multi-class. 

The type of numerical sample imbalance may also vary. Step imbalance has a cliff effect (e.g., some classes have 1,000 examples, others only 10), whereas exponential imbalance involves a graduated difference in the number of samples in a multi-class setting.

The central problem of imbalanced data is that ML models under-perform when predicting the label of minority classes. Under-performance is usually measured in a variety of ways to account for the numerical imbalance of class samples. Balanced accuracy (BAC) equally weights individual class accuracy. Macro F1 equally weights precision and recall for the classes. Area under the curve (AUC) is also used \cite{sokolova2009systematic}.

The reason why ML classifiers have difficulty generalizing to classes with fewer instances has been attributed to: numerical class imbalance \cite{krawczyk2016learning}, class overlap \cite{batista2004study,denil2010overlap,prati2004class,garcia2007empirical}, sub-concepts, and disjuncts \cite{jo2004class,weiss2004mining}. Minority class prediction under-performance can manifest itself in two different ways. In the case of single-layer SVM and LG models, classifiers may under-fit the minority class and exhibit low minority class accuracy. In the case of over-parameterized CNNs, they can over-fit the minority class, with high training, but low test accuracy \cite{dablain2022understanding}. 

In the case of under-fitting, a natural remedy is to force the classifier to pay more attention to the minority class by increasing the cost of misclassification. This can be done indirectly by increasing the number of minority samples (through DA or over-sampling) or directly, by algorithmically increasing the penalty. Over-fitting is usually addressed the same way, however, the goal is slightly different. Instead of forcing the model to \textit{attend} to the minority, the goal is to encourage the model to recognize a greater number and diversity of minority real or latent features when making a prediction. This process usually involves regularizing model parameters. 

The leading methods used to address imbalanced data are: resampling \cite{Koziarski:2021}, cost-sensitive algorithms \cite{Siers:2021}, ensemble methods \cite{Wegier:2022}, and decoupling feature representations from classification \cite{Huang:2016,Khan:2018}.  Resampling equalizes the number of majority and minority class samples by under-sampling the majority and over-sampling the minority. In this paper, we focus on over-sampling, which is a form of DA, since it augments the number of minority class samples used during training.

Three leading over-sampling methods are: Synthetic Minority Over-Sampling (SMOTE) \cite{chawla2002smote}, ADASYN \cite{he2008adasyn}, and REMIX \cite{bellinger2020remix}. REMIX is significant because it incorporates the widely used mix-up \cite{zhang2017mixup} DA technique used in balanced image training to imbalanced learning. All three methods create synthetic examples to augment the minority class through interpolation of instance features, although each method draws samples from the training set in different ways. In addition to these methods, two recent DA methods for imbalance are DeepSMOTE \cite{dablain2022deepsmote}, which adapts SMOTE to deep learning and EOS \cite{dablain2022efficient}, which generates minority samples based on nearest adversary class instances.

Each of the DA techniques described above has \textit{three principal components: a sampling strategy, a feature manipulation strategy, and a label preservation strategy}. In the case of SMOTE, it randomly samples from the minority class; ADASYN samples hard to classify instances from the minority class; and REMIX both randomly under- and over-samples from the majority and minority classes. All of these methods sample at the \textit{front-end} of models in \textit{input} space. DeepSMOTE randomly samples from the minority class; and EOS randomly samples minority class and nearest class adversaries; however, both methods sample in \textit{latent} space. 

All of the methods use a form of interpolation to \textit{manipulate} instance \textit{features}. In addition, all of the methods, except for REMIX and EOS, ensure that they sample from the \textit{vicinity} \cite{chapelle2000vicinal} of features of the intended class (the minority) by searching in the space of \textit{same-class} (minority) samples. By sampling only from same-class instances and interpolating between them to create synthetic instances, the methods effectively preserve training set labels in the synthetically created data. 

Since REMIX samples from \textit{both} minority and majority instances and interpolates between them to create synthetic instances, it \textit{preserves labels} (i.e., it ensures that synthetic features are close in the vicinity of the original training set labels and features) through label smoothing \cite{szegedy2016rethinking,muller2019does}. EOS preserves labels by selecting non-same class samples that are close in distance (on the decision boundary) between the minority class and nearest adversary classes.

\subsection{Theories on why DA is so effective at improving generalization}

Despite the wide-spread use of data augmentation in imbalanced learning, few studies have been conducted to understand why or how it works \cite{elreedy2023theoretical}.  Elreedy and Atiya analyzed the distribution characteristics of synthetic data generated by SMOTE, how the augmented data deviated from the original distribution, and developed a mathematical analysis of the covariance matrix of SMOTE generated data \cite{elreedy2019comprehensive}. Goodman et al. categorize synthetic DA generation techniques for minority classes into broad categories based on their sampling strategy \cite{goodman2022distance}. They identify: (1) methods that sample based on proximity of minority class instances, (2) methods that sample based on density and their distance from majority classes, and (3) methods that attempt to model the underlying distribution of the minority class.

Outside of imbalanced learning, several theories have discussed why DA successfully improves generalization. Wu et al. state that DA adds new information to model training \cite{wu2020generalization}. Chen et al. propose that augmented data leads to the minimization of an augmented loss \cite{chen2020group}. Dao et al. propose that DA improves generalization by simultaneously reducing model complexity and inducing invariance \cite{dao2019kernel}.

Starting with Bishop \cite{bishop2006pattern}, another line of research focuses on the regularization effects of DA. Hernandez-Garcia et al. \cite{hernandez2018data} and Carrantino et al. \cite{carratino2020mixup} point to explicit and implicit forms of regularization in ML models. Explicit regularization includes: $L^p$ penalties on learnable parameters \cite{tikhonov1963solution,bishop1995training}, weight decay \cite{hanson1988comparing}, and drop-out \cite{srivastava2014dropout}; whereas DA is a form of implicit regularization. 

A recent line of research posits that DA improves generalization because it manipulates features. Shen et al. propose that DA forces models to focus on low frequency features in data instead of spurious, high frequency features \cite{shen2022data}. Their study is inspired by Yin et al.'s Fourier perspective on DA \cite{yin2019fourier}. Ilse et al. state that DA reduces the spurious correlation between labels and features \cite{ilse2021selecting}. Allen-Zhu et al. propose that DA purifies model weights by reducing dense mixtures of hidden parameters that focus on spurious features \cite{allen2022feature}. 

We are inspired by these works and seek to understand how DA influences model weights, support vectors and feature selection when training with imbalanced data. We focus on a particular form of synthetic data generation that is widely used in imbalanced learning that manipulates features through interpolation, which was originally developed by Chawla et al.  \cite{chawla2002smote} and then extended to many other DA algorithms \cite{kovacs2019empirical,bellinger2020remix,he2008adasyn}. As described more fully below, we select from this family a number of DA techniques that use different sampling strategies to preserve labels.

\section{Methods}
\label{sec:meth}

In order to better understand how synthetic DA works in the face of data imbalance, we consider four research questions (RQs). Our RQs, in turn, inform the methods that we design to explore these RQs. All of the RQs are framed in terms of imbalanced data.

Our RQs can be summarized as: 

\begin{itemize}
    
    \item {RQ1:}  What has a greater effect on model performance: class numerical equalization or feature manipulation?
    
    \item {RQ2:} Is the choice of where to augment (real or latent space) affected by data type?
    
    \item {RQ3:} How does DA for imbalance affect model weights and support vectors?
    
    \item {RQ4:} How does synthetic DA, in real or latent space, affect front-end feature selection?

\end{itemize}

We now discuss the methods employed in pursuit of the RQs.

\subsection{Where to augment: extension of EOS framework to SMOTE \& tabular data.}

Because we would like to assess whether it is more impactful to augment in model latent vs. real space, we extend the latent DA method of EOS to SMOTE. This provides us with two methods, which both use interpolative feature manipulation, but different instance sampling strategies, that operate in \textit{latent} space. 

In EOS, the authors outlined a framework for the implementation of nearest adversary over-sampling in a deep network's latent space before the final classification layer. We extend this approach to SMOTE, a popular over-sampling technique \cite{chawla2002smote}. SMOTE has historically been used as a pre-processing step at the front-end of single layer ML models. In DeepSMOTE \cite{dablain2022deepsmote}, the authors incorporated SMOTE over-sampling into deep networks through the use of an auto-encoder trained with a reconstruction plus a penalty loss. 

This approach necessarily required training two models: an auto-encoder and a classifier. It also relied upon front-end data augmentation. In other words, for highly imbalanced datasets, it led to the creation of a large number of high-dimensional images that consumed valuable GPU training time. A frequent criticism of imbalanced learning over-sampling methods, as compared to cost-sensitive methods, is their costly increase in training time.

To address these issues, we propose a simplification and modification to DeepSMOTE. Instead of using a separate auto-encoder to generate data to train a model; we use the model itself to learn an encoding of the data and then over-sample in latent space. The over-sampled data is used to re-train a classifier, much like EOS, except we use same-class sampling instead of nearest adversary sampling. Henceforward, we refer to the modified version of DeepSMOTE as DSM to avoid confusion.
 
We apply DSM and EOS, which perform DA in latent space, to both image and tabular data and compare them with DA techniques that work at the front-end of model learning.

\subsection{Counting and comparing support vectors (SV)}

An under-studied aspect of DA techniques concerns how they affect the selection of support vectors in SVM models. SVM models operate differently than LG or CNNs because they store a subset of training \textit{instances}, whereas LG and CNN models use input \textit{features} to train weights. At inference time, SVMs determine the similarity of SVs with test data. The similarity of a test instance to each SV is weighted by a dual coefficient (Lagrange multiplier). 

In order to better understand the impact of DA on SVs, we track the identity and class label of each SV, which corresponds to an instance in the original training set. This allows us to compare the number of SVs with and without DA and whether SVs represent synthetic data instances or original training data.

\subsection{Measurement of weight norms \& differences in model parameters}

In contrast to SVMs, at inference time, LG and CNN models apply learned weights to input \textit{features}. Therefore, in order to better understand the impact of DA on learned weights in LG, NN and CNN models, we examine weight norms and differences in weight amplitudes.

We use the Frobenius norm ($\mathcal{F}$) to provide a scalar value of weight vectors \cite{golub2013matrix}. The Frobenius norm can be expressed as:

\begin{equation}
  \mathcal{F} = \sqrt{\Sigma_i,_j(w_i,_j)^2}
\end{equation}

where $w$ are model weights.

The norm allows us to compare the relative sizes of majority and minority weights to determine if models favor (over-weight) the features of one class vs. another. It also permits us to determine if DA has a regularization effect, which generally causes a reduction in the cumulative magnitude of model weights.

In addition, we calculate differences in model weights. For this purpose, we compare the absolute value of the difference of weights in models trained with imbalanced data (base models) with those in models trained with DA. We then divide the difference by the absolute value of base model weights to determine the percentage of change. These computations provide us with an indication of whether DA causes material changes in learned model parameters on a percentage basis.

\subsection{Collection of salient pixel \& tabular indices to identify changes in front-end feature selection}

In order to assess how DA affects model feature selection, we focus on the most important features. Feature importance can be determined through a variety of techniques, including local interpretable model explanations (LIME) \cite{ribeiro2016should} and Shapley values  \cite{shapley1997value,sundararajan2020many}. These techniques focus on feature importance for \textit{single instances} instead of \textit{entire classes or datasets}. They can also be computationally expensive because they involve repeated forward passes through a model \cite{achtibat2022towards}. We instead adopt a top-K approach, which we tailor to LG, CNN and NN models. We apply this approach to determine how DA changes feature selection for models trained with imbalanced vs. augmented data.

\RestyleAlgo{ruled}
\begin{algorithm}[h]
\scriptsize

\caption{LG Models: Top-K CE}\label{alg:ch8_CE}

\DontPrintSemicolon 
\BlankLine
$W = weights$\\
$F = features$\\
$top-K = number\ of\ top\ features$\\
$C = class$\\
$N = number\ of\ classes$\\
$B = Baseline\ model\ trained\ with\ imbalanced\ data$\\
$Models = LG\ models\ trained\ with\ B\ or\ augmented\ data$\\


\BlankLine
\For {$each\ M$}{
\For {$each\ C$}{
$CE = F\ X\ W$\;
$CE_{mu} = mean(CE)$\;
$CE_{arg} = argsort(CE_{mu})$\;
$M_{top-K} = CE_{arg}[:top-K]$\;
$Store\ M_{top-K}$
 } }

\For {$each\ M\ (except\ B)$}{
\For {$each\ C$}{
$I_C = intersection\ of\ M_{top-K}\ \&\ B_{top-K}$\;
$I_M = \frac{\Sigma I_C}{(N\ X\ top-K)}$\;
 } }

\BlankLine
\BlankLine
\vspace{-.2cm}
\end{algorithm}

For tabular data used in single layer LG models, there is a direct alignment of feature indices used in the classification step and input feature indices. Here, we collect \textit{classification embeddings (CE)}. CE represent the output of learned weights multiplied by input features, before they are summed. During the classification step, LG models aggregate all CE through summation and then make a prediction based on the summation of CE magnitudes. For example, in binary classification with the Sigmoid function, a CE summation which is negatively signed will produce a class 0 prediction, whereas a CE summation that is positively signed will produce a class 1 prediction.

\RestyleAlgo{ruled}
\begin{algorithm}[h]
\scriptsize

\caption{NN \& CNN Models: Top-K Input Gradients}\label{alg:ch8_NN}

\DontPrintSemicolon 
\BlankLine
$G = model\ gradient\ with\ repsect\ to input$\\
$I = test\ set\ instance$\\
$top-K = number\ of\ top\ features$\\
$N = number\ of\ instances(I)$\\
$B = Baseline\ model\ trained\ with\ imbalanced\ data$\\
$Models = models\ trained\ with\ B\ or\ augmented\ data$\\


\BlankLine
\For {$each\ M$}{
\For {$each\ I$}{
$IG_{arg} = argsort(G)$\;
$MIG_{top-K} = IG_{arg}[:top-K]$\;
$Store\ MIG_{top-K}$\;
\If {M==B}{
$Store\  MIG_{top-K}\ as\ BIG_{top-K}$\;}
 } }

\For {$each\ M\ (except\ B)$}{
\For {$each\ I$}{
$I_C = intersection\ of\ MIG_{top-K}\ \&\ BIG_{top-K}$\;
$I_M = \frac{\Sigma I_C}{(N\ X\ top-K)}$\;
 } }

\BlankLine
\BlankLine
\vspace{-.2cm}
\end{algorithm}

For each instance, the absolute value of each individual CE magnitude can be extracted from a LG model. Each of these CE magnitudes has a corresponding index. The magnitudes and indices can be sorted to determine the indices that contributed the most to the model's decision for an individual instance. The indices, in turn, can be aggregated for all instances and the ones that occur most frequently (the top-K) can be determined. See Algorithm~\ref{alg:ch8_CE}. This approach is more fully described in \cite{dablain2022understanding}.

In contrast, for NN and CNN models, there is no direct correspondence between CE and input features. The dimensionality of the CE encoding is typically compressed compared with the input feature dimensionality. Therefore, we back-propagate the model's label prediction to determine the gradient of each input feature using the method of \cite{simonyan2013deep}. Here, we heuristically select K-percent of the input features for a single instance that have the highest gradient. 

For tabular data, where the location (index in an input vector) of features does not change, we aggregate all feature indices for each instance and determine the feature (indices) that occur most frequently in a dataset or class. In the case of image data, the location of a feature may change due to scale, pose, differences in class objects, etc. Therefore, we make comparisons on an instance by instance basis.  We compare the percentage of common features between models trained with imbalanced data and augmented data. For each instance, we determine a percentage of the top-K that are common to the same instance, but determined with two different models (the intersection of two top-K sets). Because the instances of imbalanced and augmented training datasets vary, we use the test sets. Then, we average the intersection percentage across all instances. See Algorithm~\ref{alg:ch8_NN}.

We use this approach to determine whether DA changes the most salient or important features that a model uses to make its predictions in dense neural network and CNN models. 

\section{Experimental set-up}

\subsection{RQ1: class numerical equalization vs. feature manipulation}
\label{ch8_rq1_exp}

To gain insight into RQ1, we examine two broad types of imbalanced learning methods: (1) those that \textit{only} balance the classes and (2) those that numerically balance classes \textit{and} incorporate some form of feature interpolation. For the first group, we select simple random over-sampling (ROS). We also select a simple cost-sensitive (CS) method that weights each class based on the relative number of samples to provide a basic comparison between DA and CS methods. Both ROS and CS reweighting are designed to equalize the classes - one by increasing the number of minority class training samples and the other by increasing the cost of incorrect label predictions during supervised training. In contrast, DA methods that use interpolation not only balance the classes, but they also form synthetic instances by varying input features. To simply our analysis, we do not consider under-sampling methods.

We conduct our experiments with two types of data: tabular and image. For the tabular data, we select SMOTE, ADASYN and REMIX as our DA methods and use three different types of parametric classifiers: SVM \cite{vapnik1999nature}, LG, and a dense, 4-layer neural network (NN). The NN is structured with an input layer dimension that varies with the number of input features, followed by layers with 100 and 50 neurons. The final two layers contain: (1) neurons tailored to half the input size, which varies by dataset, and then (2) two neurons to perform binary classification. 

For the image data, we select REMIX, DSM and EOS as our DA methods and use a CNN as our classifier \cite{fukushima1982neocognitron,lecun1998gradient}. We use a standard Resnet architecture for the CNN \cite{he2016deep}. We use the same training regime for all methods, as employed by Cao. et al. \cite{cao2019learning}. We also report the results of a cost-sensitive algorithm, LDAM \cite{cao2019learning}, which is widely used as a reference algorithm in imbalanced image training, to provide context for our DA methods. For this purpose, we implement the version of LDAM without delayed re-weighting (DRW) to make it comparable to our reference DA methods, which also do not use delayed re-weighting.

The tabular and image datasets used in this experiment are described below.

\subsubsection{Description of tabular datasets}

We select five binary classification datasets for testing from the UCI machine learning library \cite{Dua:2019}. The datasets are: Ozone, Scene, Coil, Thyroid and US Crime. Our dataset selection criteria are: imbalance ratio greater than 10:1, number of samples greater than 1,000, number of features greater than 50, and datasets containing only non-categorical features (i.e., integer and real numbers). The key details of the datasets are summarized in Table~\ref{tab: ch8_data}.

\begin{table}[h!]
\vspace{-.2cm}
\footnotesize
\caption{\textbf{Tabular Datasets}}
\label{tab: ch8_data}
\centering
\begin{tabular}{ p{1.3cm}p{.7cm}p{1cm}p{1cm}}
\toprule

\textbf{Dataset} & \textbf{Imb. Ratio} &
\textbf{\# \mbox{Samples}} &
\textbf{\# \mbox{Features}} \\

\midrule

Ozone & 34:1 & 2,536 & 72 \\
Scene & 13:1 & 2,407 & 294 \\
Coil & 16:1 & 9,822 & 85 \\
Throid & 15:1 & 3,772 & 52 \\
US-crime & 12:1 & 1,994 & 100 \\
\bottomrule

\end{tabular}
\vspace{-.2cm}
\end{table}

For each dataset, we perform 5-way cross-validation, with a 70:30 split between training and test sets. All datasets were scaled using zero mean and unit variance. We report balanced accuracy and macro F1 measure based on averaging over five splits of the data.  

We use these tabular datasets in all of our subsequent RQs.

\subsubsection{Description of image datasets}

For image data, we select three datasets: CIFAR-10 \cite{krizhevsky2009learning}, Places \cite{zhou2017places}, and INaturalist \cite{van2018inaturalist}. CIFAR-10 is initially balanced and we imbalance it exponentially, with a 100:1 maximum level. For the Places dataset, we select five classes: airfield, amusement park, acquarium, baseball field and barn. Each class is initially balanced. We employ 10:1 step imbalance (airfield and amusement park with 5K training samples and the rest with 500). From the INaturalist dataset, we randomly select samples from five classes: plant, insect, bird, reptile and mammal, with 5:1 step imbalance. 

For each dataset, we select three different random combinations of initially larger sized training classes. For CIFAR-10, we use the same test set to evaluate each training sample because the test data is drawn from a different sample space than the training data. For Places and INaturalist, we randomly select different training and test set samples from separate training and validation sets that do not overlap. Table~\ref{tab:ch8_img_data} contains details of the image training and test sets. We report our results averaged over three different random combinations of the training sets.

\begin{table}[h!]
\vspace{-.1cm}
\footnotesize
\caption{\textbf{Image Datasets}}
\label{tab:ch8_img_data}
\centering
\begin{tabular}{ p{1.2cm}p{.5cm}p{.5cm}p{.7cm}
p{.7cm}p{.7cm}p{.5cm}p{.5cm}}
\toprule

\textbf{Dataset} & \textbf{\# Class} &
\textbf{Imbal. Type} &
\textbf{Max \mbox{Imbal.} Level} & \textbf{\# Train Maj.} &
\textbf{\# Train Min.} &
\textbf{Num. \mbox{Test} \mbox{Maj.}} &
\textbf{\mbox{Num.} Test \mbox{Min.}}\\

\midrule

CIFAR-10 & 10 & Expon. & 100:1 & 5000 & 50 & 1000 & 1000\\

Places & 5 & Step & 5:1 & 2500 & 500 & 250 & 250\\

INaturalist & 5 & Step & 5:1 & 6250 & 1250 & 500 & 250\\

\bottomrule

\end{tabular}
\vspace{-.1cm}
\end{table}

We use these same image datasets in our subsequent RQs.


\subsection{RQ2: where to augment}
\label{ch8_rq2_exp}
For RQ2, we compare DA algorithms implemented in input space (SMOTE, ADASYN, REMIX) with ones that are implemented in latent space (DSM and EOS). For tabular data, we use a NN and for image data, we use a CNN. For image data, we use REMIX in real space, since it is designed for image data, whereas SMOTE and ADASYN are not. 

\subsection{RQ3 \& RQ4}
For these RQs, we use the trained models and data from the first two experiments and compare changes in models trained with imbalanced data with those trained with augmented data. We measure differences in weights, support vectors and feature selection based on the methods described in Section~\ref{sec:meth}.

\section{Results}

\subsection{RQ1: numerical equalization vs. feature manipulation}

\begin{figure}[h!]
   \vspace{-.2cm}
   \footnotesize
  \centering
  \subfloat[SVM: BAC]{\includegraphics[width=0.24\textwidth]{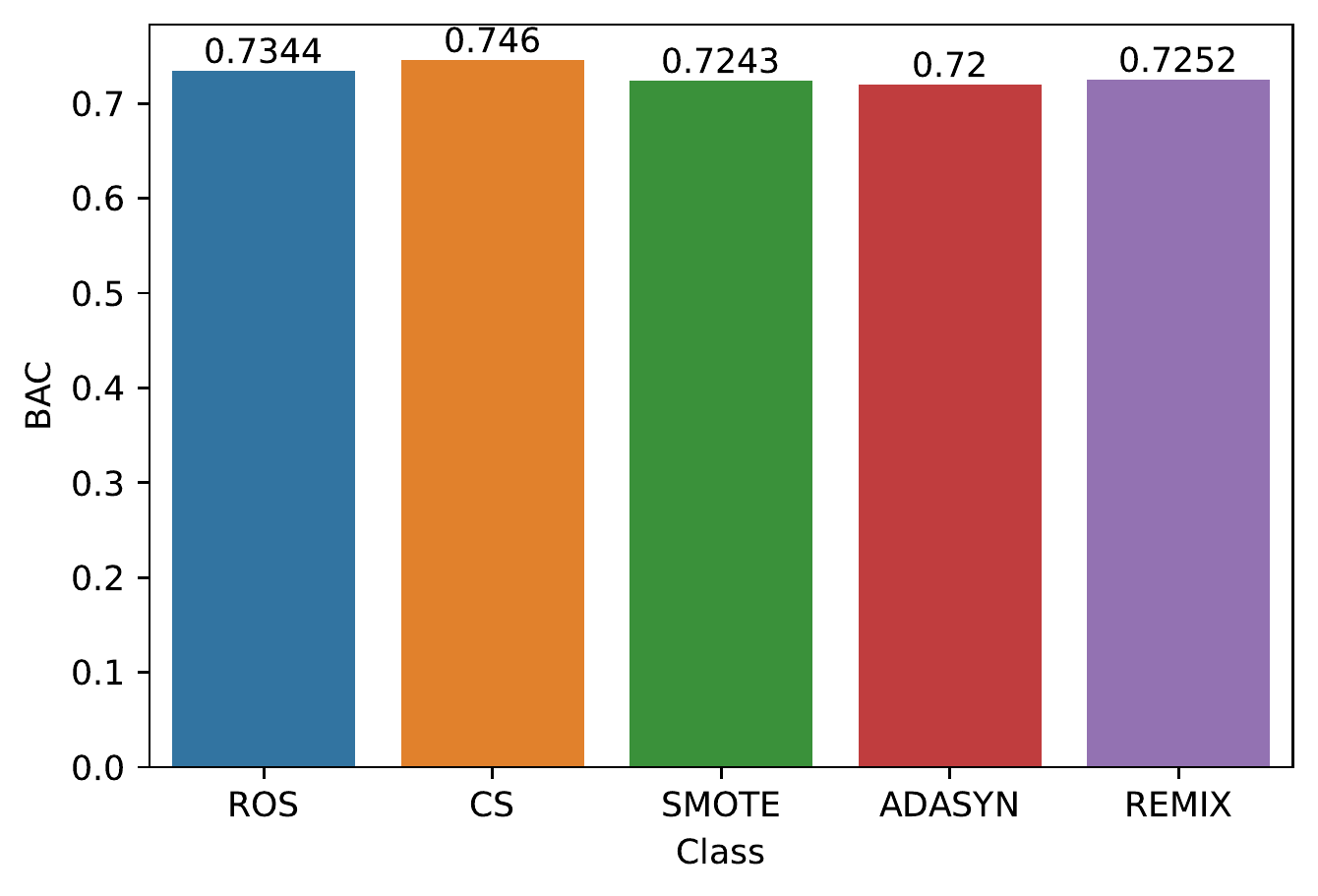}}
  \hfill
  \subfloat[SVM: FM]{\includegraphics[width=0.24\textwidth]{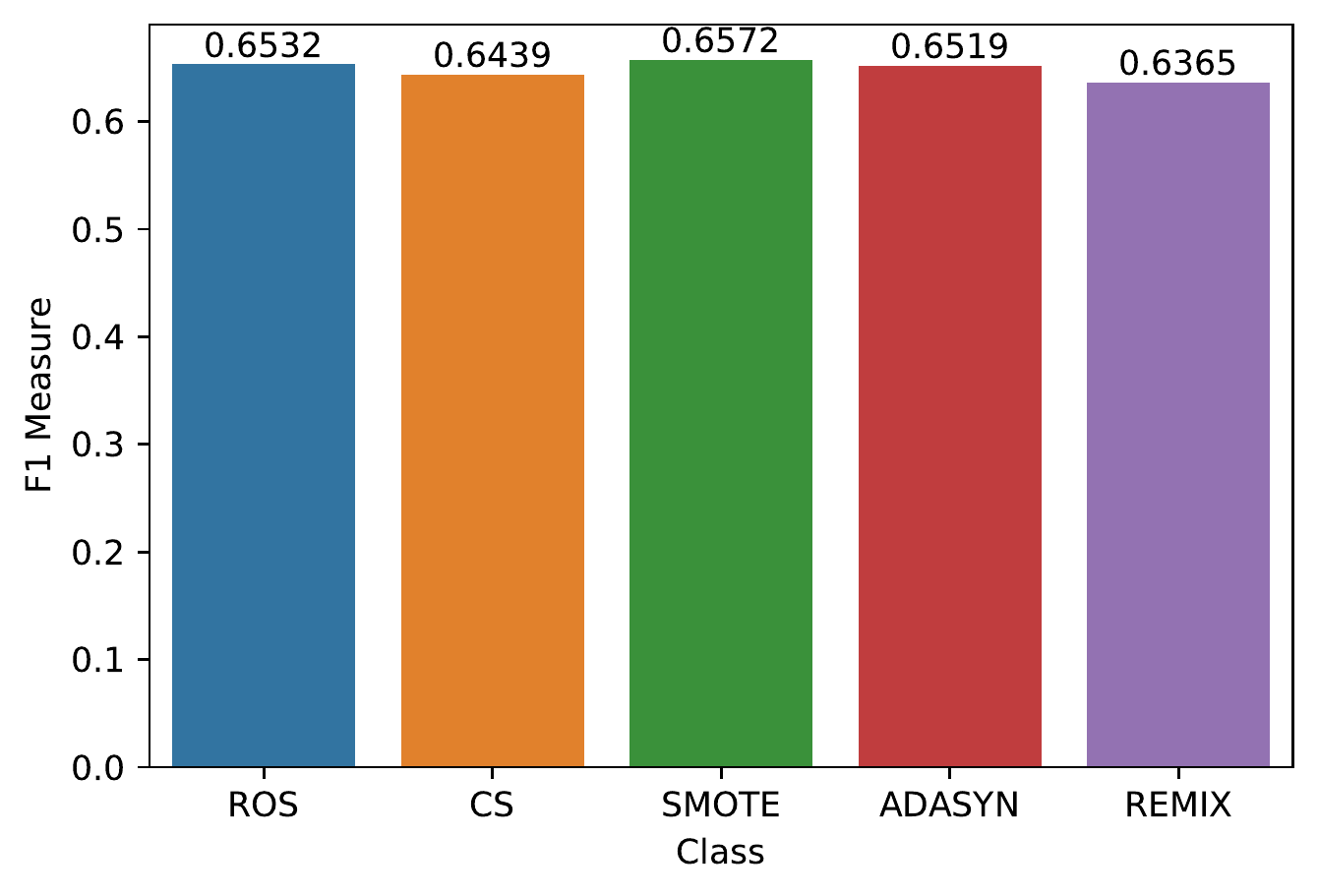}}
  \hfill
  \subfloat[LG: BAC]{\includegraphics[width=0.24\textwidth]{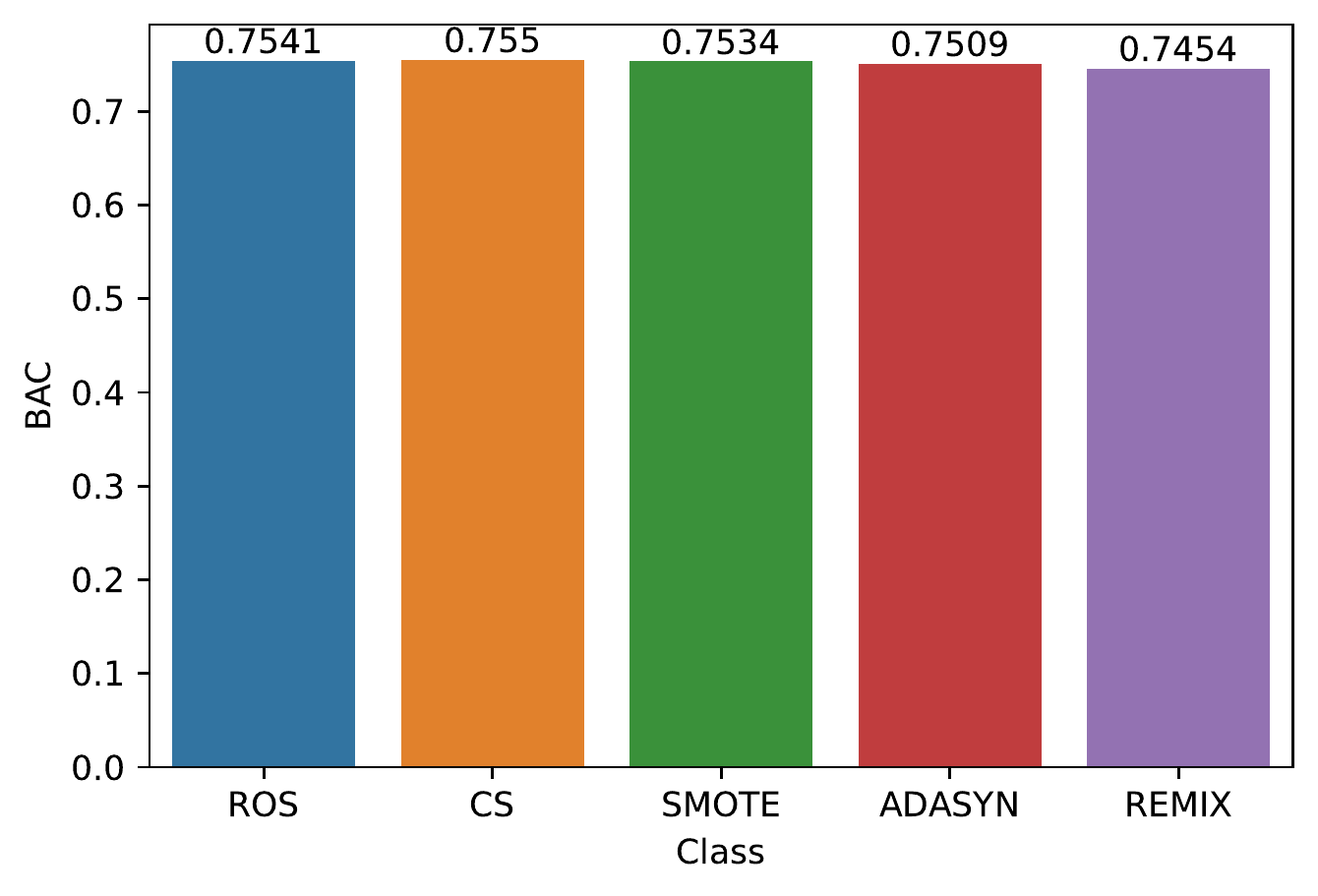}}
  \hfill
  \subfloat[LG: FM]{\includegraphics[width=0.24\textwidth]{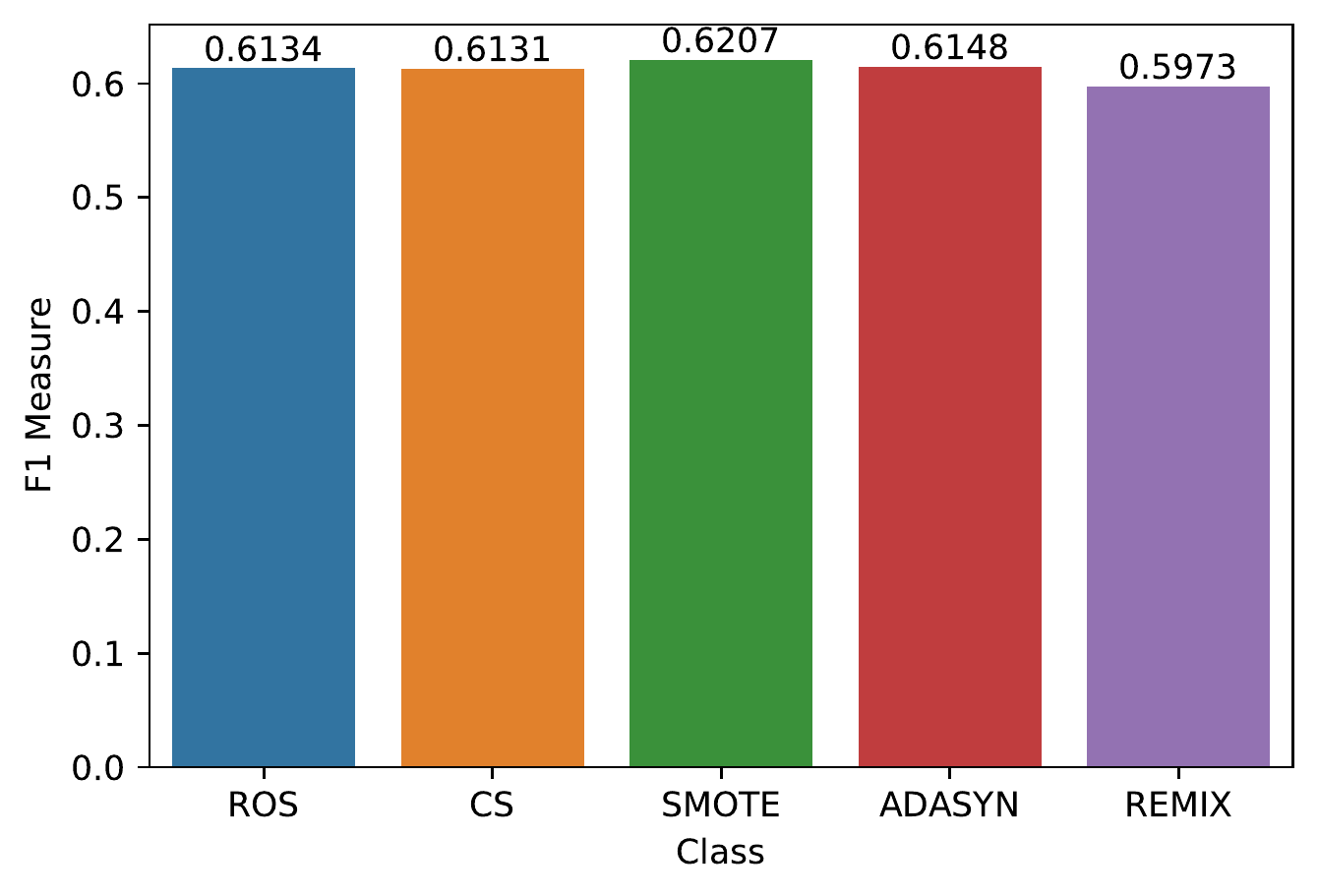}}
  \hfill
  \subfloat[NN: BAC]{\includegraphics[width=0.24\textwidth]{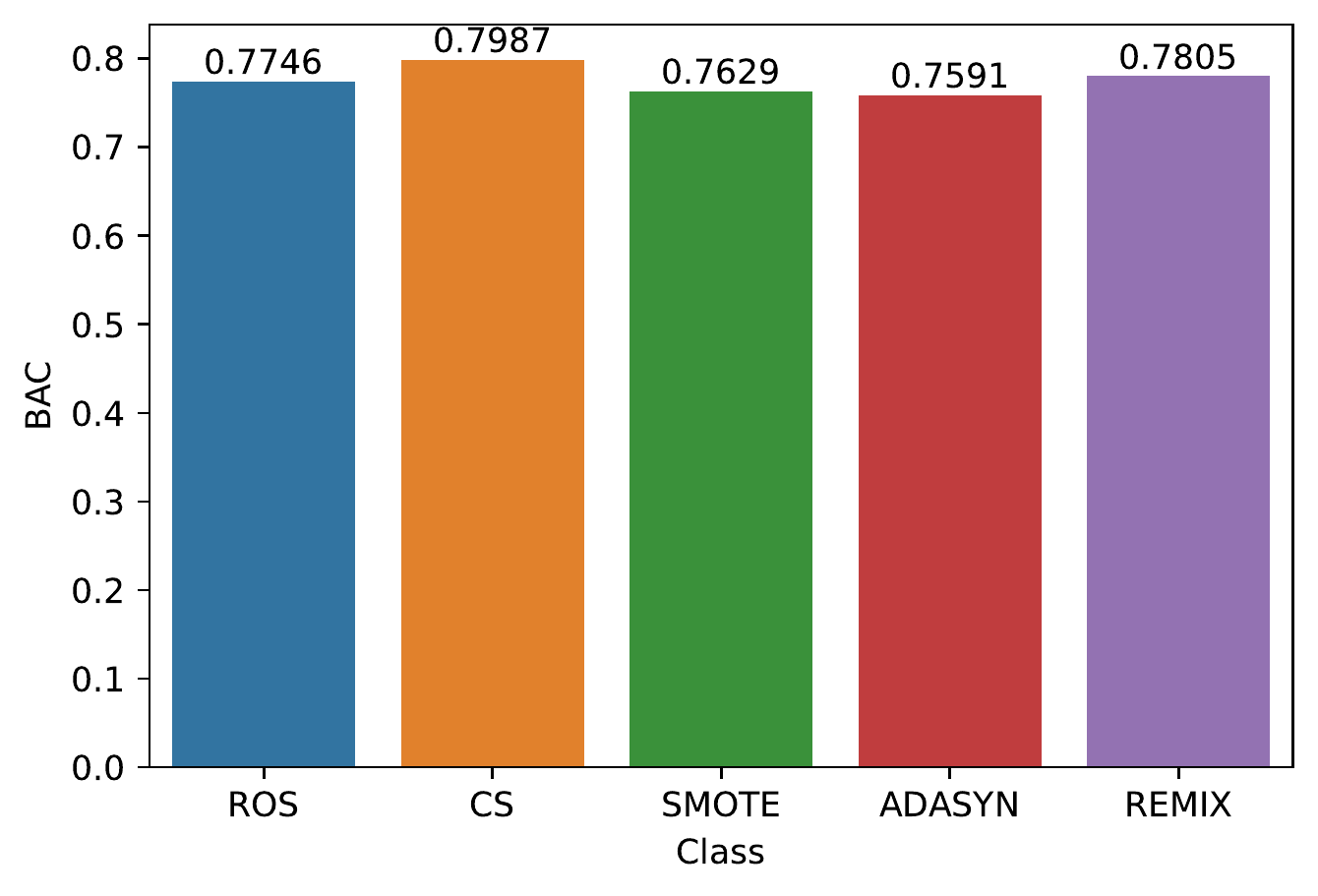}}
  \hfill
  \subfloat[NN: FM]{\includegraphics[width=0.24\textwidth]{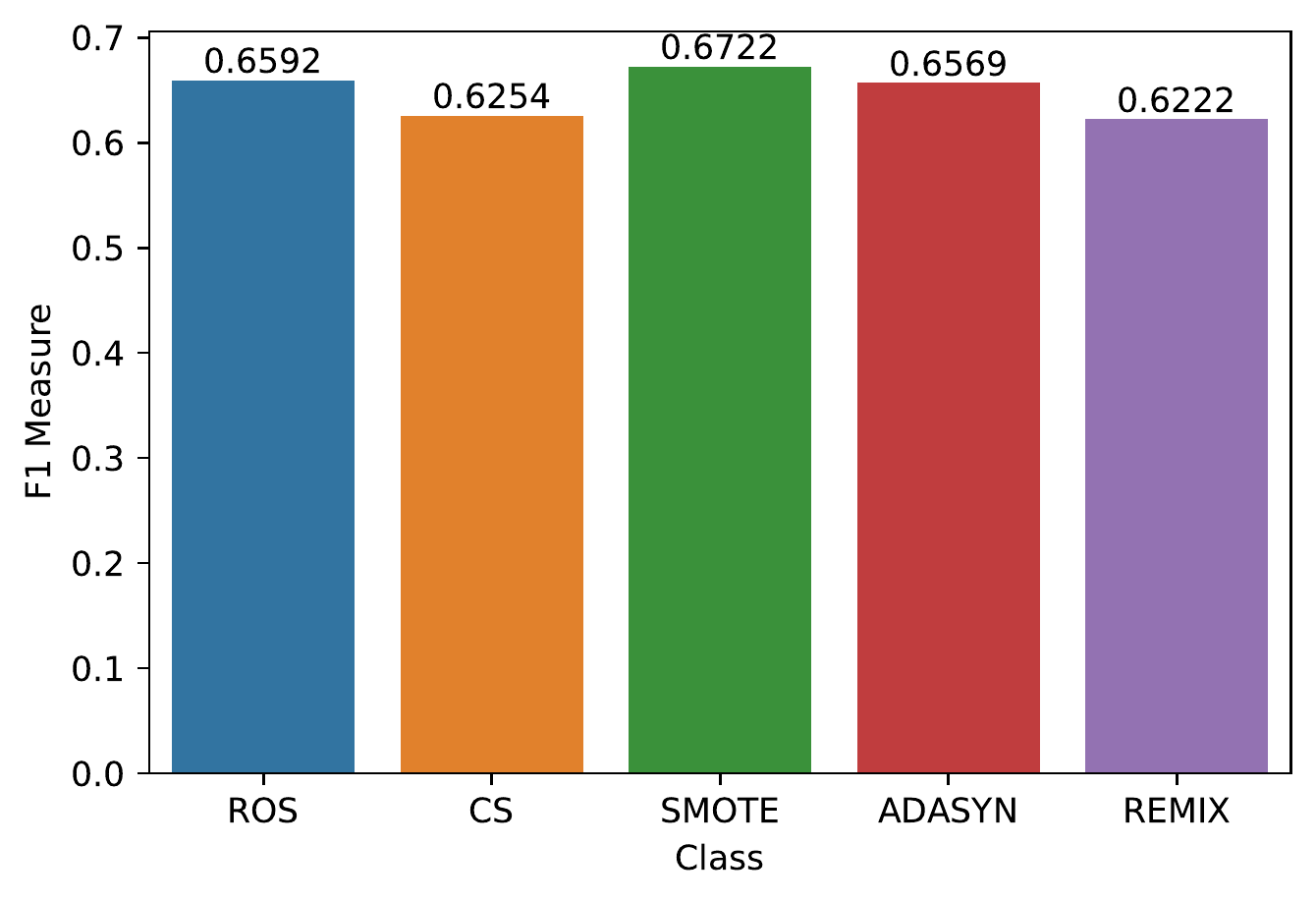}}

  \caption{These charts compare the BAC and macro F1 of methods that numerically equalize the number of classes (ROS and CS) with DA methods that use numerical balancing plus feature manipulation (SMOTE, ADASYN and REMIX). Results for 3 different classifiers (SVM, LG and NN) are shown, averaged over 5 datasets and 5 cross-validation runs.}
  \label{fig_ch_8_perf1}
  \vspace{-.2cm}
\end{figure}

Figure~\ref{fig_ch_8_perf1} shows the results of imbalanced over-sampling and cost-sensitive re-weighting methods on tabular data. The results include BAC and macro F1 averaged over five cross-validation runs for each method and datasets based on three classifiers (SVM, LG and NN). For purposes of these figures, we contrast the relative performance of methods that merely seek to balance the classes (ROS and CS) vs. methods that perform class rebalancing plus feature interpolation (SMOTE, ADASYN and REMIX). Table~\ref{tab:Ch8_tab_results} provides a more detailed listing of the performance of each method by dataset.

In the case of LG and SVM classifiers, there are marginal differences between methods that perform numerical balancing of classes vs. interpolative DA methods based on BAC and FM. The differences in BAC and FM are approx. $\mp 1$. In the case of NN, the differences are $\mp 3$; however, they vary by DA method. For example, there is less than a $\mp 1.5$ difference between ROS and the same-class interpolative methods (SMOTE and ADASYN); with a greater difference between CS and the DA methods.

\begin{table}[h!]

\centering
\footnotesize
\vspace{-.1cm}
\caption{Detailed Performance Results for Tabular Data}
\label{tab:Ch8_tab_results}
\begin{tabular}{p{1.5cm}|p{.6cm}p{.7cm}|
p{.6cm}p{.7cm}| p{.6cm}p{.6cm}}
\toprule

\multicolumn{1}{l}{\textbf{}} & 
\multicolumn{2}{c}{\textbf{SVM}} & 
\multicolumn{2}{c}{\textbf{LG}} &
\multicolumn{2}{c}{\textbf{NN}} 
\\
\midrule

Method & BAC & FM & BAC & FM & BAC & FM \\

\midrule

\multicolumn{7}{l}{\textbf{Ozone}}\\
\midrule

Base & .5000 & .4829  & .5471 & .5609 &
 .7160 & \textbf{.6534}   \\

ROS & .7567 & \textbf{.6363}  & .7929 & .5878 & .7950 & .6067  \\

CS & \textbf{.7738} & .6159  & .7956 & .5859 & \textbf{.8316} & .6040  \\

SMOTE & .7499 & .6293  & \textbf{.8004} & \textbf{.5956} & .7986 & .6137  \\

ADASYN & .7507 & .6317 & .7994 & .5936 &
 .7989 & .5954  \\

REMIX & .7522 & .6157 & .7782 & .5578 &
 .8178 & .5990  \\

\midrule

\multicolumn{7}{l}{\textbf{Scene}}\\
\midrule

Base & .5033 & .4874  & .5754 & \textbf{.5803} & .6469 & .6357   \\

ROS & .6080 & .6227  & .6381 & .5714 &
 .6721 & .6083  \\

CS & \textbf{.6404} & .6229  & \textbf{.6441} & .5742 &
 \textbf{.7204} & .5552  \\

SMOTE & .6041 & .6223  & .6361 & .5683 &
 .6547 & \textbf{.6344} \\

ADASYN & .6087 & .6282  & .6358 & .5699 &
 .6530 & .6224  \\

REMIX & .6014 & .6171  & .6221 & .5515 &
 .6847 & .5952 \\

\midrule

\multicolumn{7}{l}{\textbf{Coil}}\\

\midrule
Base & .5000 & .4847 &  .5067 & .4989 &
 .5090 & .5042   \\

ROS & .6137 & .5339  & .6681 & .5051 &
 .6411 & .5429  \\

CS & \textbf{.6255} & .5317  & \textbf{.6686} & .5042 &
 \textbf{.6695} & .5153  \\

SMOTE & .5963 & .5378 & .6600 &
.5019  & .6116 & \textbf{.5491}  \\

ADASYN & .5906 & .5336  & .6593 & .4993 &
 .6110 & .5371  \\

REMIX &  .6083 & \textbf{.5437}  & .6614 &
\textbf{.5086} & .6607 & .5312  \\

\midrule

\multicolumn{7}{l}{\textbf{Thyroid}}\\

\midrule
Base & .5080 & .5005  & .7645 & \textbf{.8068} & .8595 & .8290   \\

ROS & .8806 & .7331  & .8696 & .6965 &
 \textbf{.9251} & .8220 \\

CS & .8784 & .7209  & .8655 & .6959 &
 .9191 & .7647  \\

SMOTE & \textbf{.8875} & \textbf{.7648} &  .8725 & .7243  & .9165 & \textbf{.8325}  \\

ADASYN & .8830 & .7457  & \textbf{.8759} & .7117  & .9206 & .8100  \\

REMIX & .8490 & .6986  & .8695 & .6810  & .8952 & .7387  \\

\midrule
\multicolumn{7}{l}{\textbf{US Crime}}\\

\midrule
Base & .5942 & .6327  & .7100 & \textbf{.7426}  & .7882 & \textbf{.7581}\\

ROS & .8129 & \textbf{.7401}  & \textbf{.8017} & .7063  &
 .8399 & .7162  \\

CS & .8121 & .7283  & .8013 & .7051 &
 \textbf{.8529} & .6876  \\

SMOTE & .7836 & .7317  & .7980 & .7134 &
 .8331 & .7314  \\

ADASYN & .7669 & .7201  & .7839 & .6994 &
 .8120 & .7197  \\

REMIX & \textbf{.8152} & .7077  & .7961 & .6878 & .8438 & .6467  \\

\bottomrule

\end{tabular}
\end{table}

Based on the information contained in Table~\ref{tab:Ch8_tab_results}, the average \textit{baseline} BAC and macro FM for all 5 datasets is .6153 and .6105, respectively, across all classifiers. By simply equalizing the number, or cost of incorrect minority class predictions,  average BAC increases to .7544 and .7666 for ROS and CS, respectively, and average macro FM increases to .6419 and .6275, respectively. If we average the performance of the interpolative DA methods (SMOTE, ADASYN and REMIX), their BAC and FM is .7469 and .6366, respectively. 

Thus, DA with feature manipulation does not appear to provide substantial performance enhancement over simple class numerical rebalacing for these \textit{tabular} datasets, when measured in terms of BAC and FM. This lack of improvement may be partially attributable to the documented problems of SMOTE-based algorithms with higher dimensionality datasets (here, dimensions of only 50 to 300) \cite{blagus2013smote}. The problem has been diagnosed as an issue with the nearest neighbor algorithm used by SMOTE-based algorithms. However, in our experiments, we include REMIX, which does not use nearest neighbors to select combinatorial instances. 

In contrast, Table~\ref{tab:ch8_latent_res} paints a different, though somewhat uneven, picture in the case of feature manipulation with image data. In the case of the CIFAR-10 and Places datasets, DA with interpolative feature manipulation with EOS and DSM outperforms numerical balancing methods (ROS and LDAM) by 3$+$ points. However, with INaturalist, ROS outperforms all other methods. Thus, for 2 out of 3 image datasets, incorporating feature manipulation outperforms simple random over-sampling. 

The potential benefits of using latent space feature interpolation with image data (EOS and DSM) are discussed in more detail in the next section.


\subsection{RQ2: DA in real vs. latent space}

Table~\ref{tab:ch8_latent_res} compares the performance of several imbalanced DA methods that are alternately implemented in the real and latent space of a NN classifier with tabular data. In the table, first place performance is bolded, second place is underlined and third place is italicized. DSM and EOS, which are implemented in latent space, perform well, with consist top 3 finishes. 

\begin{table}[h!] 

\centering
\footnotesize
\vspace{-.2cm}
\caption{Tabular Data: Latent vs. Real DA}
\label{tab:ch8_latent_res}
\begin{tabular}{ p{.3cm}p{.3cm}p{.3cm}
p{.3cm}p{.3cm}p{.3cm}p{.3cm}
p{.3cm}p{.3cm}p{.3cm}p{.3cm}
}
\toprule

\multicolumn{1}{l}{\textbf{Method}} & 
\multicolumn{2}{c}{\textbf{Ozone}} & 
\multicolumn{2}{c}{\textbf{Scene}} &
\multicolumn{2}{c}{\textbf{Coil}} &
\multicolumn{2}{c}{\textbf{Thyroid}} &
\multicolumn{2}{c}{\textbf{US Crime}} 

\\
\midrule

Method & BAC & FM 
& BAC & FM & BAC & FM &
BAC & FM & BAC & FM  
\\

\midrule


Base &  .7160 & .6534  & .6469 & .6357 & .5090 & .5042 &  .8595 & .8290  &  .7882 & .7581    \\
ROS &  .7950 & .6067   &  .6721 & .6083 &  .6411 & \textit{.5429} &  \textbf{.9251} & \underline{.8220} & .8399 & .7162 \\
CS & \textbf{.8316} & .6040  &  \textbf{.7204} & .5552 &  \textbf{.6695} & .5153 &  .9191 & .7647  &  \textbf{.8529} & .6876     \\
SMOTE & .7986 & \textit{.6137}  &  .6547 & \underline{.6344} &.6116 & \textbf{.5491} &  .9165 & \textbf{.8325} &  .8331 & \textbf{.7314}    \\
ADA. &  .7989 & .5954  &  .6530 & .6224 & .6110 & .5371 & \underline{.9206} & \textit{.8100} &  .8120 & \textit{.7197}   \\
REMIX &  \underline{.8178} & .5990  & \underline{.6847} & .5952 &  \underline{.6607} & .5312 &  .8952 & .7387 &  .8438 & .6467    \\
EOS & \textit{.8134} & \underline{.6176} & \textit{.6805} & \textit{.6316}  & \textit{.6492} & .5379 & \textit{.9194} & .7995 & \underline{.8526} & .7193  \\
DSM & .8092 & \textbf{.6286} & .6661 & \textbf{.6491}  & .6462 & \underline{.5450} & .9122 & .8078  & \textit{.8505} & \underline{.7307} \\

\bottomrule

\end{tabular}
\end{table}

However, on balance, a greater number of first place finishes are recorded with DA implemented in real space for these tabular datasets.

In contrast, Table~\ref{tab:ch8_img_res} compares the performance of several synthetic DA methods that are alternately implemented in real and latent space of a CNN classifier for imbalanced image data. In this case, DA implemented in latent space finishes first in 2 out of 3 datasets; second in all 3; and third in 1 out of 3 datasets.

\begin{table}[h!] 

\centering
\footnotesize
\vspace{-.2cm}
\caption{Image Data: Latent vs Real DA}
\label{tab:ch8_img_res}
\begin{tabular}{ p{1.6cm}p{.6cm}p{.8cm}|
p{.6cm}p{.8cm}|p{.6cm}p{.8cm}
}
\toprule

\multicolumn{1}{l}{\textbf{}} & 
\multicolumn{2}{c}{\textbf{CIFAR-10}} & 
\multicolumn{2}{c}{\textbf{INaturalist}} &
\multicolumn{2}{c}{\textbf{Places}} 

\\
\midrule

Method & BAC & FM 
& BAC & FM & BAC & FM 
\\

\midrule


Balanced &  .9265 & .9265  & .7052 & .6940 & .9016 & .7451    \\

Base &  .7232 & .7200  & .5943 & .6045 &  .7451 & .7446 \\
DSM & \underline{.7821} & \underline{.7812}  &  \underline{.6452} & \underline{.6390} &  \underline{.7912} & \underline{.7916}    \\
EOS & \textbf{.7898} & \textbf{.7889}  &  \textit{.6428} & \textit{.6356} & \textbf{.7968} & \textbf{.7971}  \\
LDAM &  .7408 & .7400  &  .5018 & .4981 & .7381 & .7401    \\
REMIX &  \textit{.7477} & \textit{.7474}  & .5788 & .5819 &  .7677 & .7697    \\
ROS & .7404 & .7394 & \textbf{.6527} & \textbf{.6529}  & \textit{.7771} & \textit{.7783}  \\

\bottomrule
\end{tabular}
\vspace{-.2cm}
\end{table}

Based on these experiments, it appears that the type of data, the depth of the network and the network architecture may affect the choice of whether to over-sample in latent vs. real space, when training with imbalanced data. Latent space DA performs well when the base model has learned a compact encoding with image data. In the case of the image data, latent space over-sampling occurs after a deep (32 or 56 layer) network has learned patterns that are present in local image patches with convolutional kernels. Conversely, with tabular data, where the initial input already consists of low-dimensional, abstracted features, DA appears to be more effective when applied at the front-end of a dense NN, when training with imbalanced data.


\subsection{RQ3: DA affect on support vectors \& weights}

\subsubsection{Support Vectors (SV)}
\begin{figure}[h!]
   \vspace{-.2cm}
  \centering
  {\includegraphics[width=0.49\textwidth]{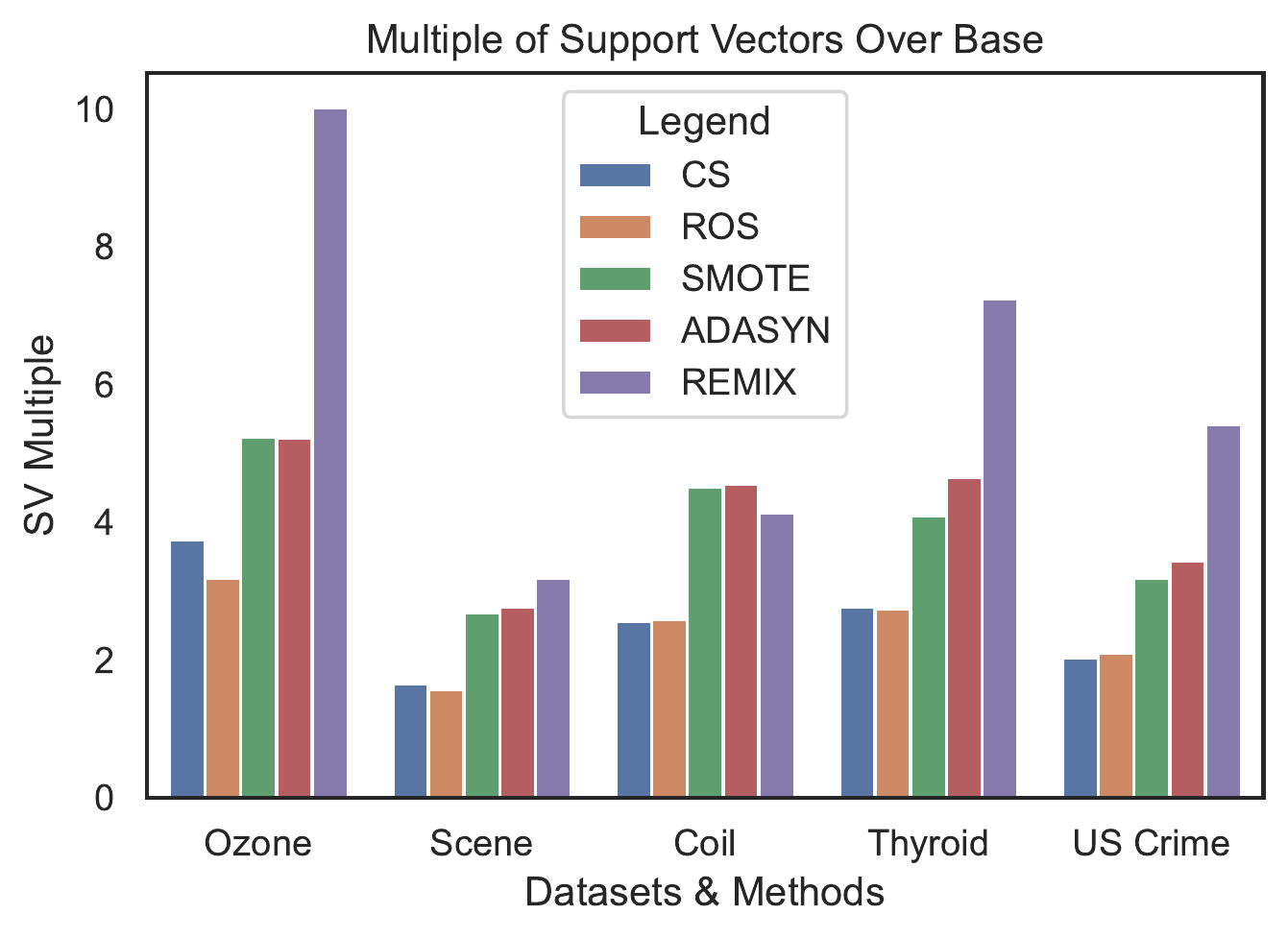}}
  \caption{This diagram demonstrates that DA consistently increases the number of support vectors over the baseline imbalanced dataset by a factor of 1 (100\%)  or greater. }
  \label{fig_ch_8_SV_mult}
  \vspace{-.2cm}
\end{figure}

In order to better understand the impact of DA on imbalanced data, we examine the number of support vectors (SVs) in SVM models trained with, and without, DA on tabular datasets. Figure~\ref{fig_ch_8_SV_mult} shows the multiple of the number of SVs for models trained \textit{with} DA and CS over a baseline model trained with imbalanced data (no DA). In all cases, DA causes the number of SVs to increase by a factor of at least 1 (100\% increase) and in some cases by much larger multiples. Models trained with DA that use feature interpolation (SMOTE, ADASYN and REMIX) show a much larger increase in the number of SVs compared to methods that do not (ROS and CS). 

This implies that models trained with interpolative DA require a larger number of SVs to disentangle classes to make predictions. We interpret this greater reliance on the number of SVs as a sign of greater \textit{memorization} of the underlying data in the face of increased noise infusion (interpolating instances instead of merely copying them). An SVM model is required to maintain in its memory a greater number of instances (or SVs) of the dataset to make predictions over a base model when it is presented with augmented data; and this number increases when DA is applied with feature manipulation, instead of simple class numerical balancing.

\begin{figure}[h!]
   \vspace{-.2cm}
  \centering
  {\includegraphics[width=0.49\textwidth]{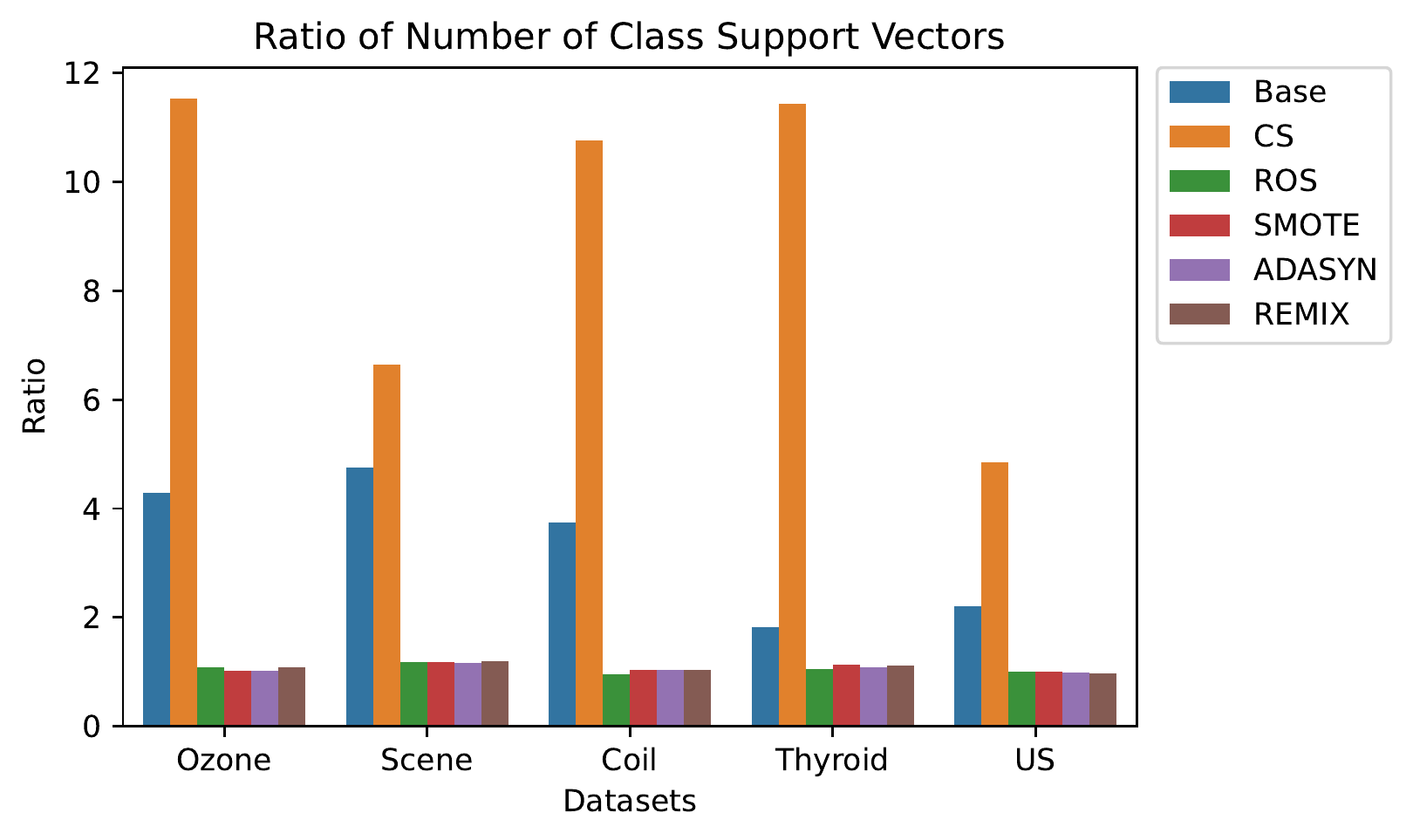}}
  \caption{This diagram shows the ratio of majority class support vectors vs minority class support vectors for each method and dataset.}
  \label{fig:sv_dcn}
  \vspace{-.2cm}
\end{figure}

Figure~\ref{fig:sv_dcn} shows the ratio of majority class to minority class support vectors for each dataset. All of the DA methods (ROS, SMOTE, ADASYN and REMIX), balance the number of class SVs compared to the base model trained with imbalanced data (i.e., the ratio of majority to minority SVs is close to 1). The base and CS models heavily favor the majority class; and yet, the CS method achieves comparable BAC and macro FM with DA methods. The CS method also uses a lower number of SVs than the interpolative DA methods (see Figure~\ref{fig_ch_8_SV_mult}).

\begin{figure}[h!]
   \vspace{-.2cm}
  \centering
  {\includegraphics[width=0.49\textwidth]{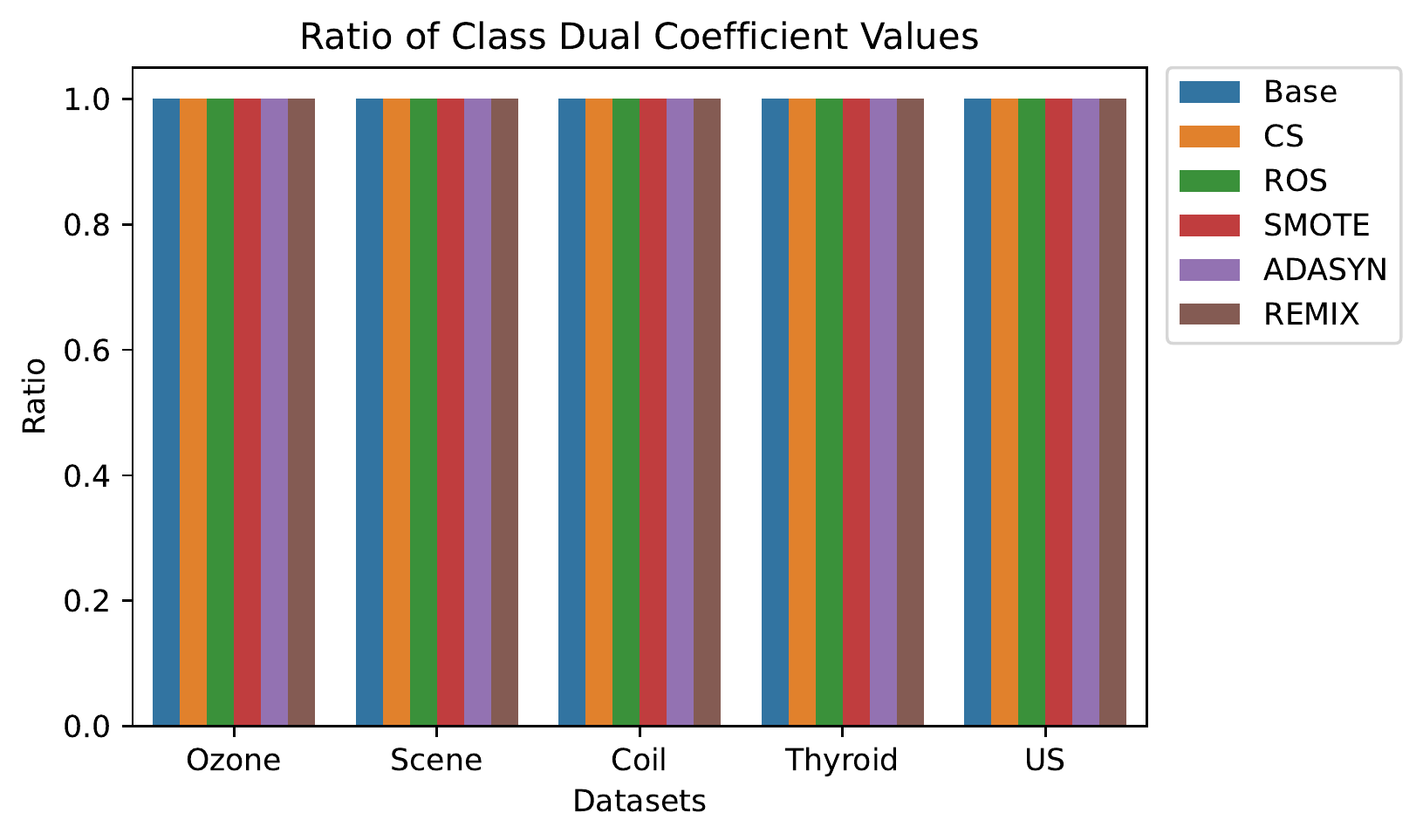}}
  \caption{This diagram shows the ratio of dual coefficients.}
  \label{fig:sv_dcv}
  \vspace{-.2cm}
\end{figure}

The reason for this apparent dichotomy emerges in Figure~\ref{fig:sv_dcv}. The aggregate value of the dual coefficients (Lagrange multipliers) for all models are equally balanced between the classes. In other words, even though there are more majority SVs for the base and CS models, the weight factors (dual coefficients) of all of the SVs for each class are equally weighted in the aggregate (i.e., the sum of dual coefficients are equal for each class, although oppositely signed).

\begin{figure}[h!]
   \vspace{-.2cm}
  \centering
  {\includegraphics[width=0.49\textwidth]{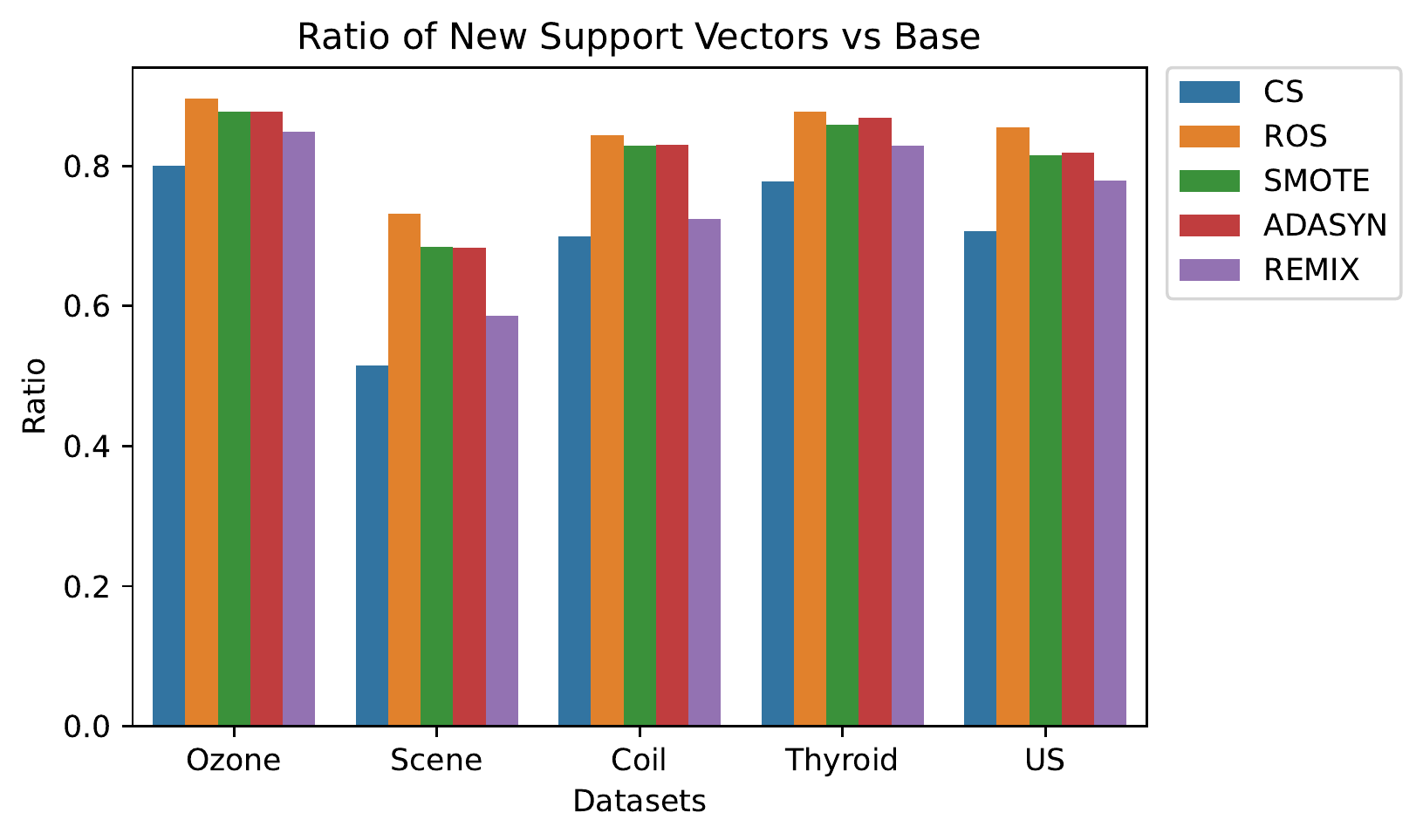}}
  \caption{This diagram shows the ratio of of support vectors in models trained with DA and CS that are not present in the base model (i.e., newly added SVs as a result of DA or CS training).}
  \label{fig:sv_new}
  \vspace{-.2cm}
\end{figure}

All models, including the base, which is trained on imbalanced data and has much lower accuracy, have balanced dual coefficients. Thus, improvements in BAC and FM are due to either a reallocation of individual dual coefficient weights among the instances and / or additional SVs (in the case of DA). 

We hypothesize that the key \textit{reason} for the improved performance when training with DA lies in the \textit{increase} in the \textit{number} of support vectors added through DA and cost-sensitive methods. Figure~\ref{fig:sv_new} shows that a large proportion of the SVs that are added by DA and CS are SVs that are \textit{not} present in the base model. In many cases, the ratio of \textit{new} SVs that are not in the Base model are over 60\% and in some cases, over 80\%. 

\begin{figure}[h!]
   \vspace{-.2cm}
  \centering
  {\includegraphics[width=0.49\textwidth]{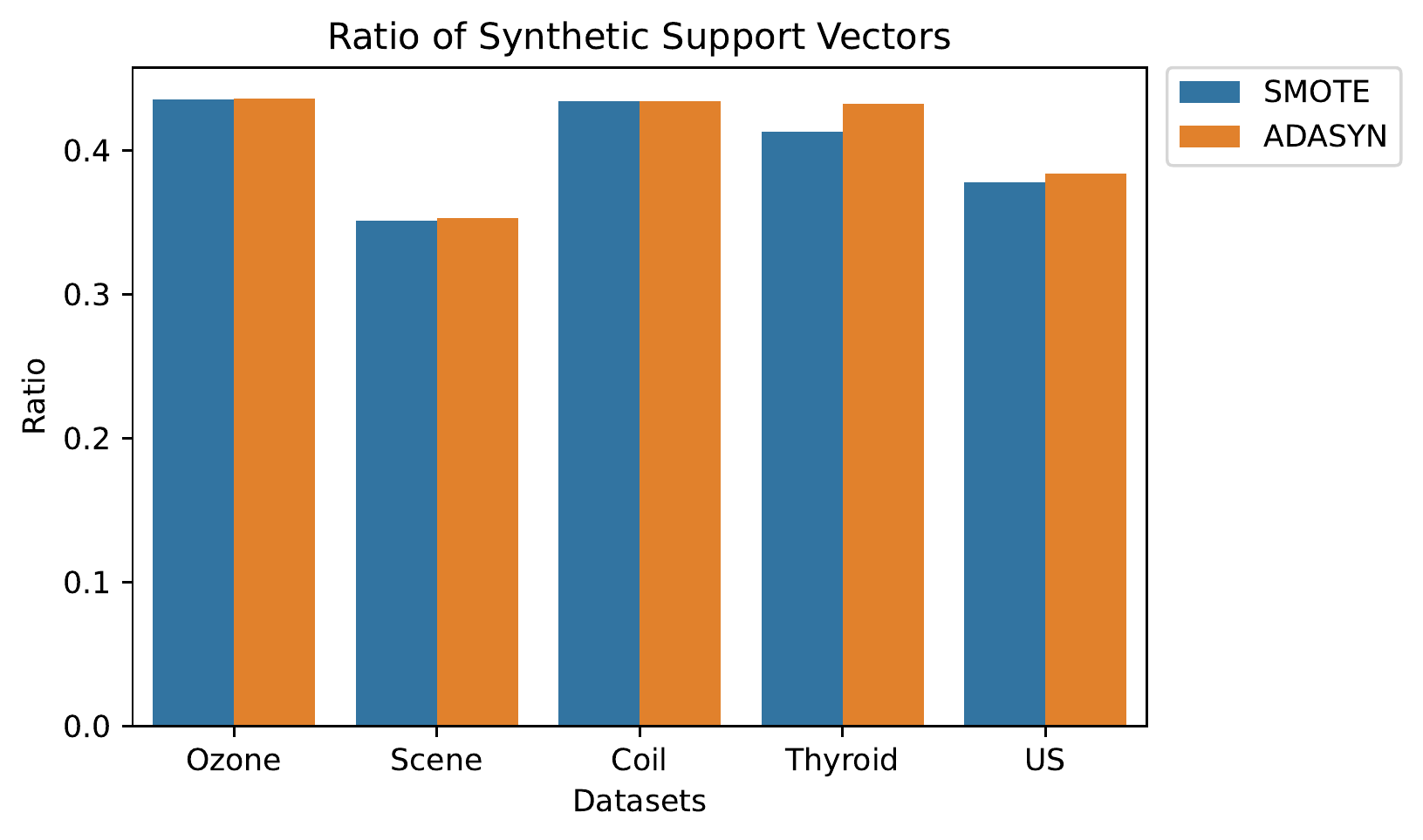}}
  \caption{This diagram shows the ratio of synthetic support vectors to all support vectors for models trained with DA (SMOTE \& ADASYN).}
  \label{fig:sv_syn}
  \vspace{-.2cm}
\end{figure}

Furthermore, Figure~\ref{fig:sv_syn} indicates that a large proportion (35\% to 40\%) of newly added SVs consist of \textit{synthetic} data, in the case of models trained with SMOTE and ADASYN.  (ROS and CS do not make use of synthetic data and REMIX both under- and over-samples the original dataset, which makes it challenging to track synthetically created instances.) 

Collectively, we believe that these results show that DA, when used with SVM models to address imbalance, works by \textit{increasing} the \textit{number} of support vectors. SVM models trained with synthetic DA, which infuse \textit{noise} into training samples through the interpolation of features, require \textit{more} SVs than models that merely reproduce the original samples without noise infusion. The fact that noisy samples require a greater number of support vectors for prediction further implies greater memorization of (noisy) instances than in the case of mere numerical balancing (ROS or CS).  

With SVM models, support vectors and dual coefficients are used to make predictions. However, LG, NN and CNN models do not use support vectors. Next, we consider the impact of DA on model weights in these models in the face of class imbalance.

\subsubsection{Weights in LG \& neural network models}

\begin{figure}[h!]
   \vspace{-.1cm}
  \centering
  \subfloat[Weight Norms]{\includegraphics[width=0.24\textwidth]{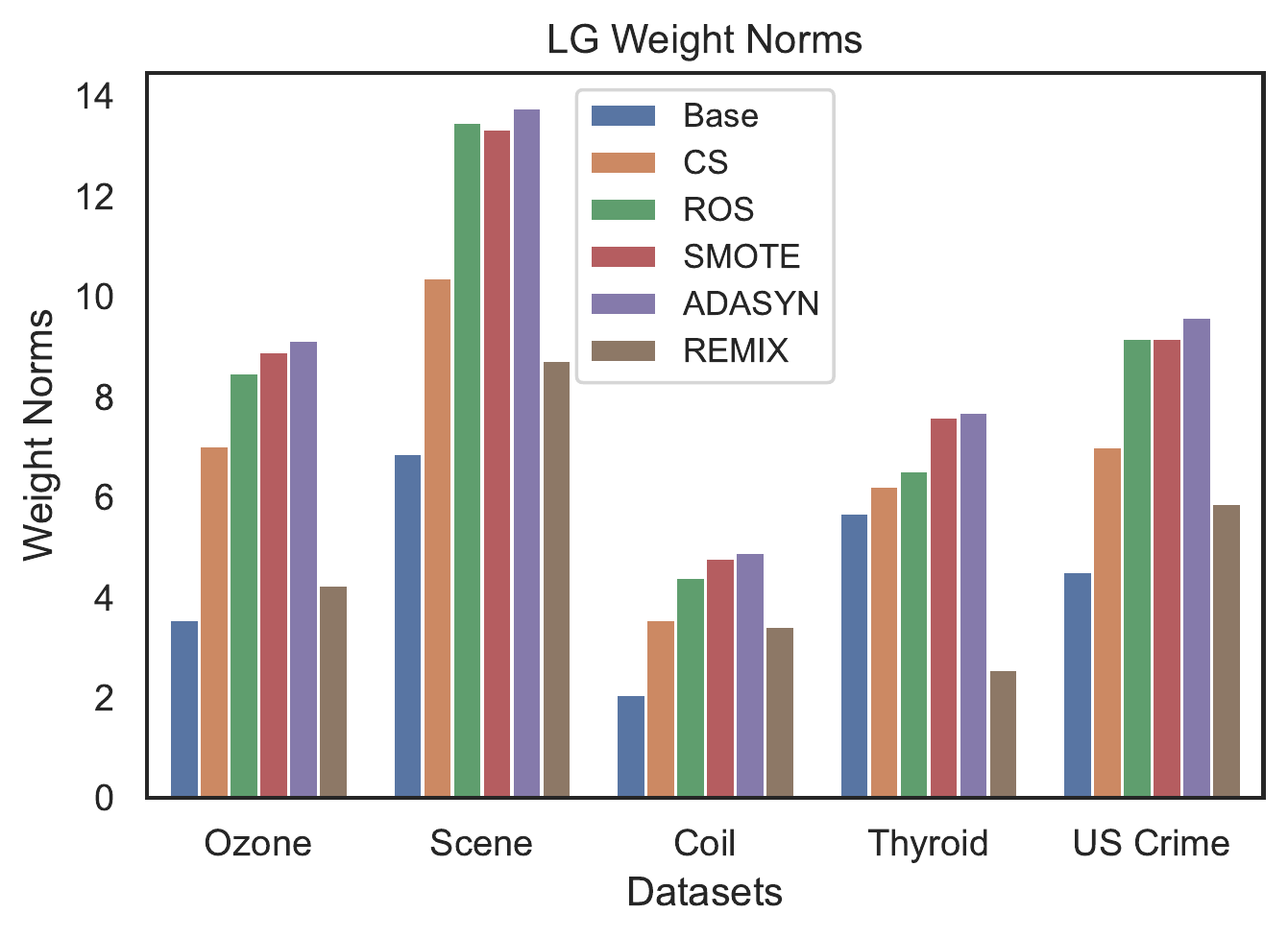}}
  \hfill
  \subfloat[Differences]{\includegraphics[width=0.24\textwidth]{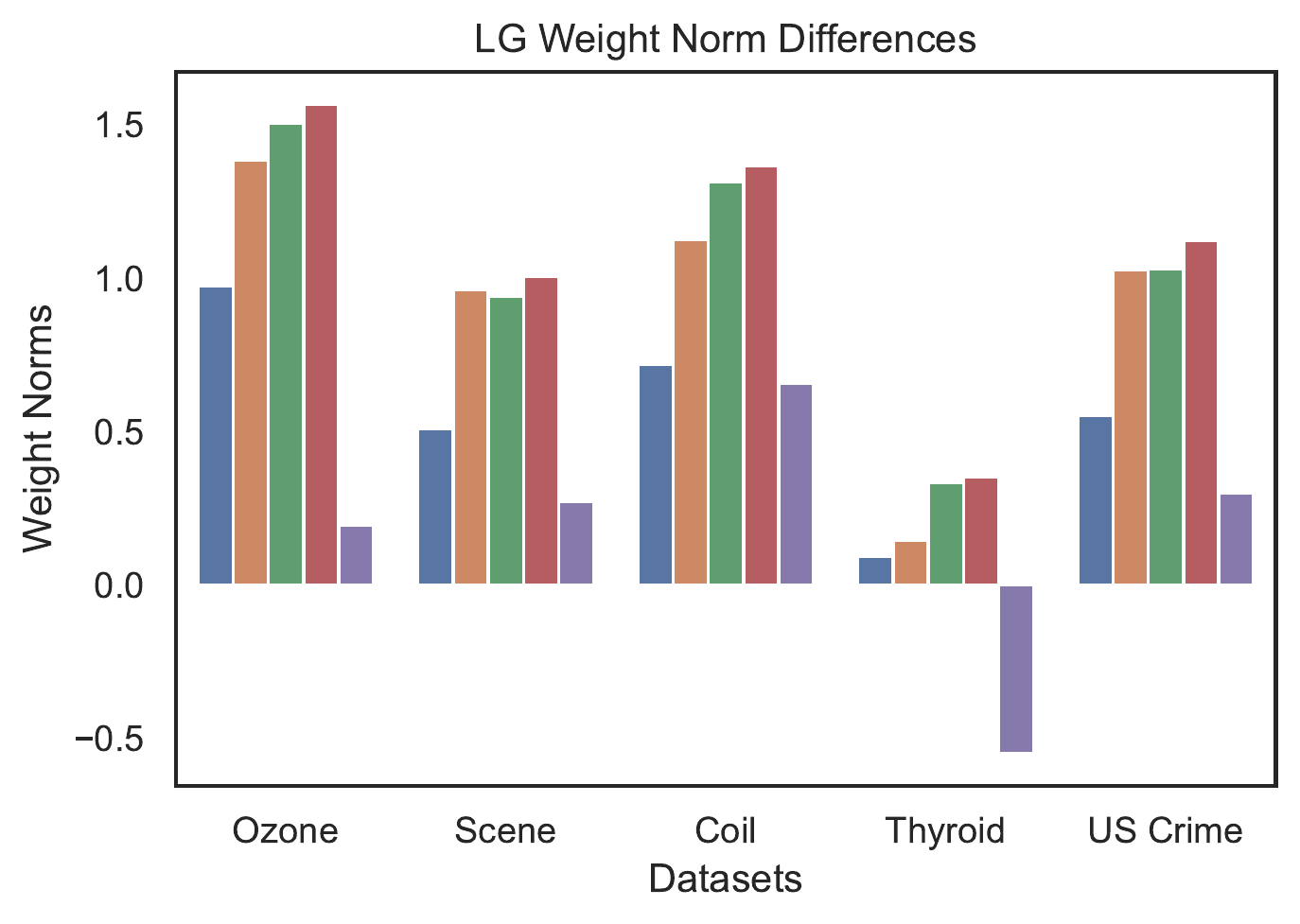}}
  \caption{This figure displays the weight norms for LG models trained with and without DA. }
  \label{fig_ch_8_LG_wts_all}
  \vspace{-.2cm}
\end{figure}

To gain an understanding of how DA affects model weights in LG, NN and CNN models, we analyze their norms. Model weights can be expressed as a vector. The norm of the weights (i.e., the square root of the summation of the square of each weight) provides an indication of the relative size of the weights. The norm can then be used to compare the relative size of weight vectors in models trained with imbalanced data and augmented data, so that we can determine if DA has a regularizing effect on model weights (i.e., it reduces their relative size). If DA does not have a regularizing effect on weights, then presumably some other factor must be responsible for the improvement of generalization with respect to minority classes.

\begin{figure}[h!]
   \vspace{-.2cm}
  \centering
 \subfloat[Majority Weight Norms]{\includegraphics[width=0.24\textwidth]{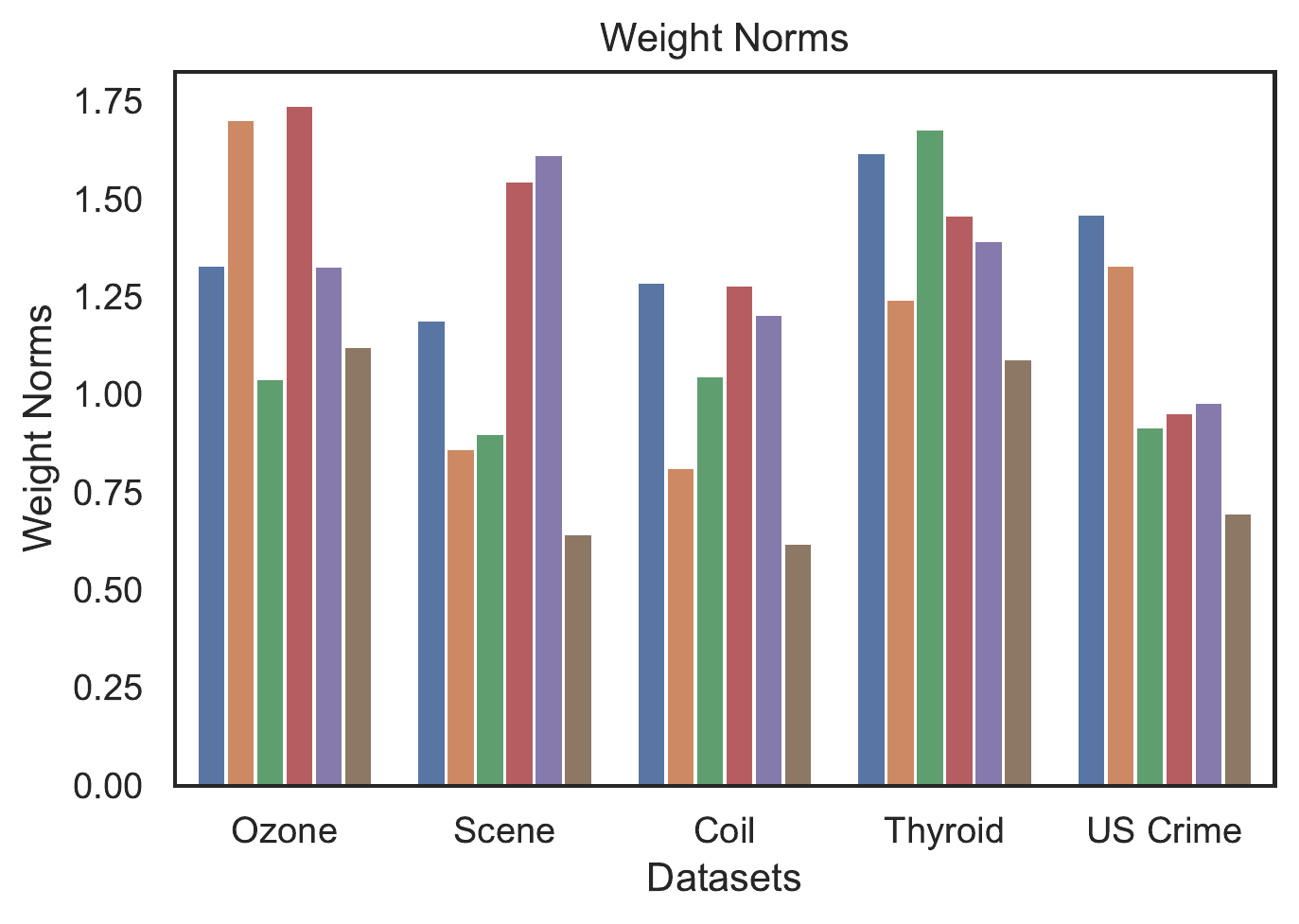}}
  \hfill
  \subfloat[Differences (Majority)]{\includegraphics[width=0.24\textwidth]{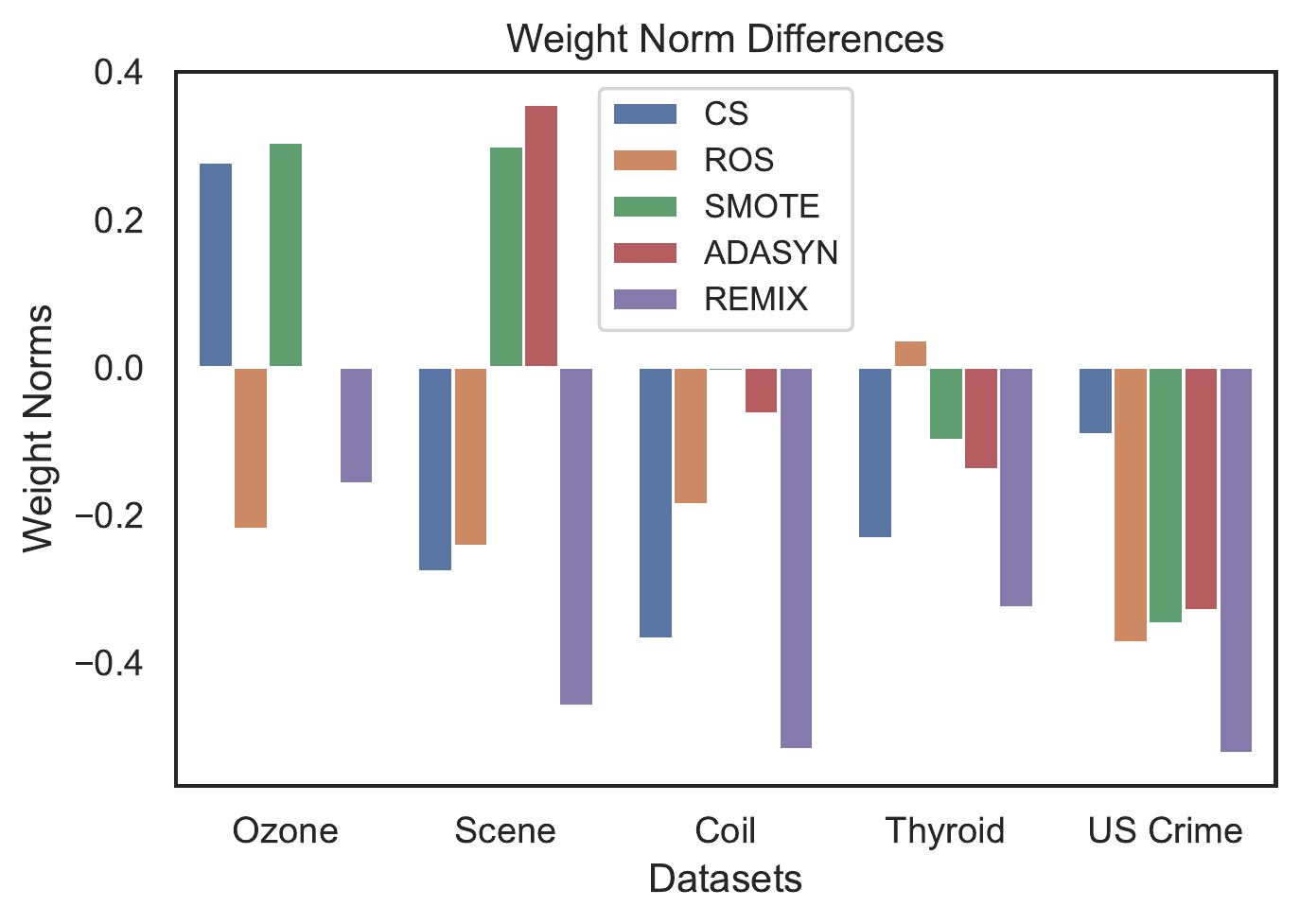}}
  \caption{This figure displays the average of all the weight norms and their differences for majority classes for NN models trained with and without DA.}
  \label{fig_ch_8_NN_wts_reg_maj}
  \vspace{-.2cm}
\end{figure}

Figure~\ref{fig_ch_8_LG_wts_all} displays the average of all LG model non-bias weight norms that were trained on imbalanced data (base), CS, and augmented data with ROS, SMOTE, ASASYN, and REMIX. 

\begin{figure}[h!]
   \vspace{-.2cm}
  \centering
   \subfloat[Minority Weight Norms]{\includegraphics[width=0.24\textwidth]{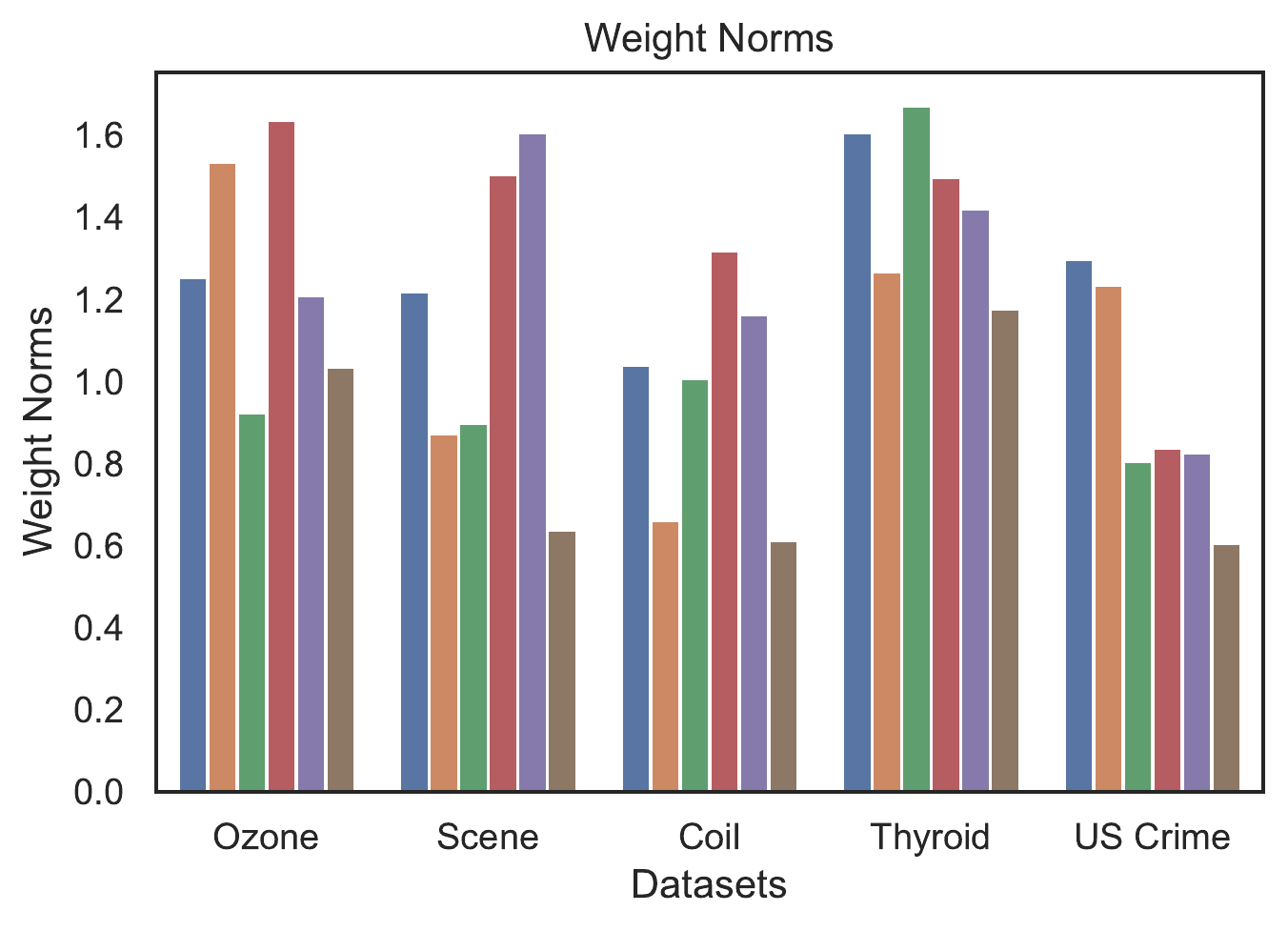}}
  \hfill
  \subfloat[Differences (Minority)]{\includegraphics[width=0.24\textwidth]{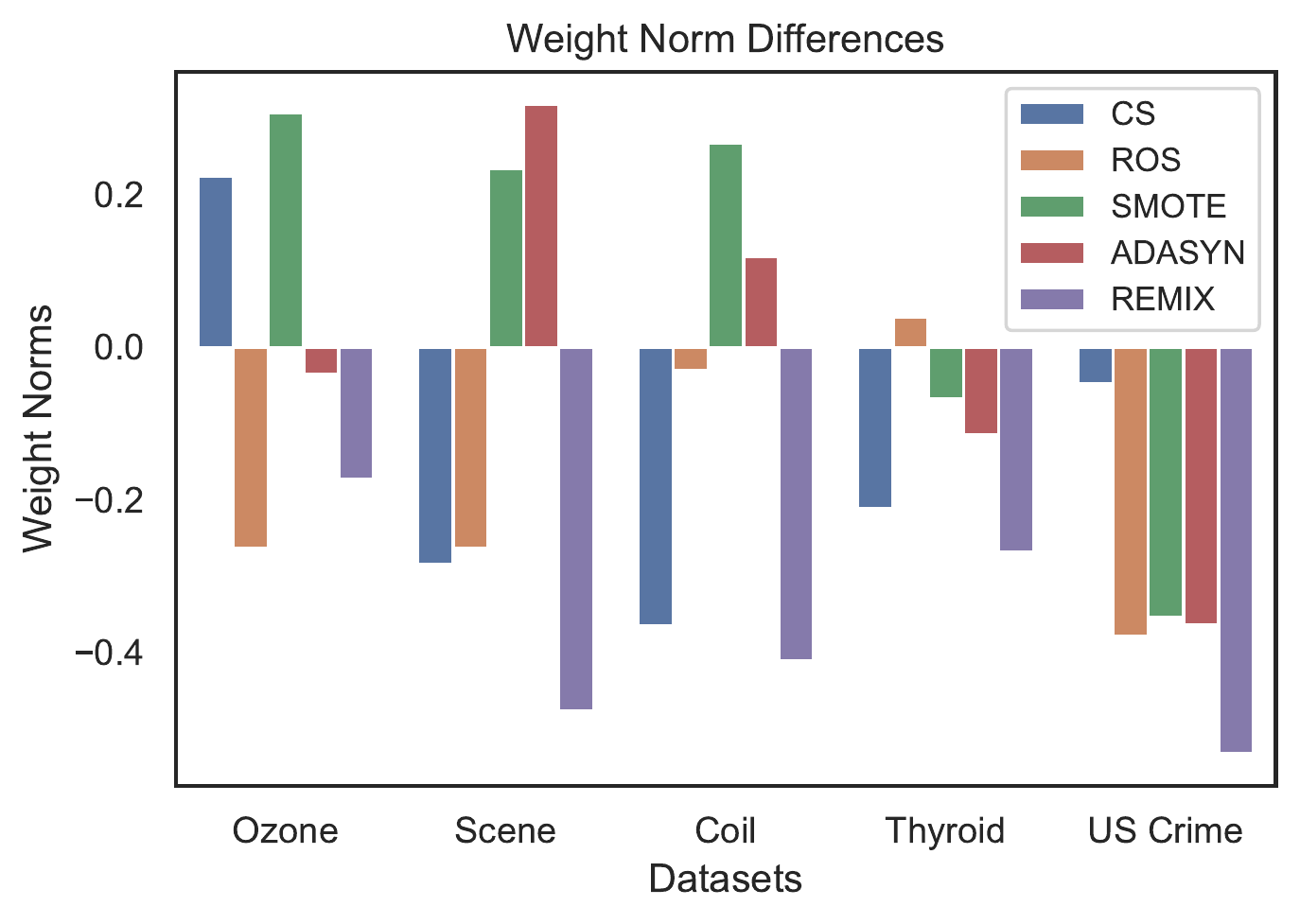}}
  \caption{This figure displays the average of all the weight norms and their differences for minority classes for NN models trained with and without DA.}
  \label{fig_ch_8_NN_wts_reg_min}
  \vspace{-.2cm}
\end{figure}

All models use L2 weight regularization. In all cases, the models trained with DA or CS resulted in an \textit{increase} in weight norms, which is seemingly the \textit{opposite} of typical regularization. The only exception is  REMIX in the case of the Thyroid dataset. For DA to have a regularizing effect, presumably it should \textit{reduce} the magnitude of most weights and increase the magnitude of a few weights, yet here the overall effect is to cumulatively increase weights.

\begin{figure*}[t!]
\vspace{-.2cm}
\centering
\subfloat[CIFAR-10]{\includegraphics[width=0.33\textwidth]{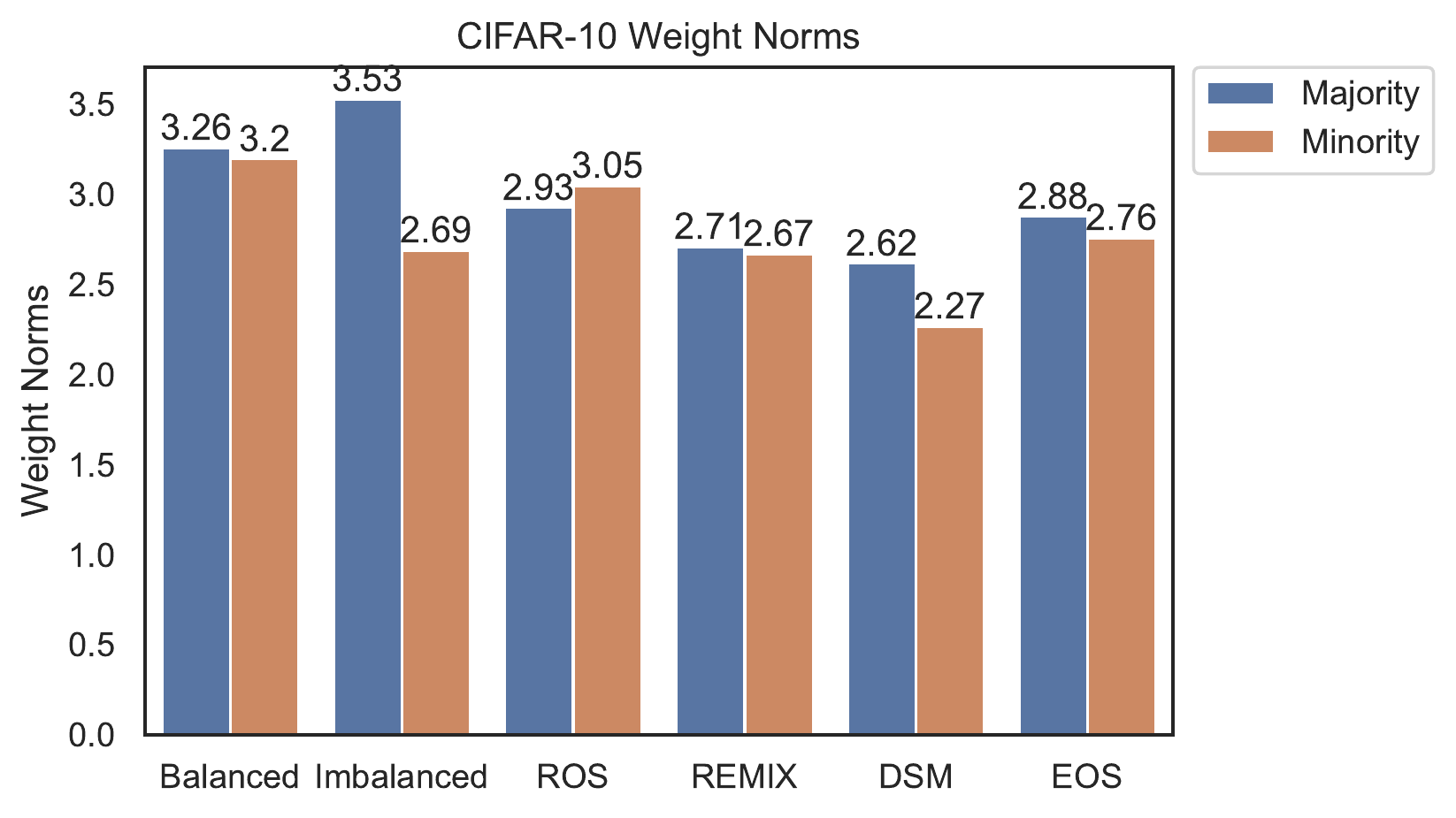}}
  \subfloat[Places]
{\includegraphics[width=0.33\textwidth]{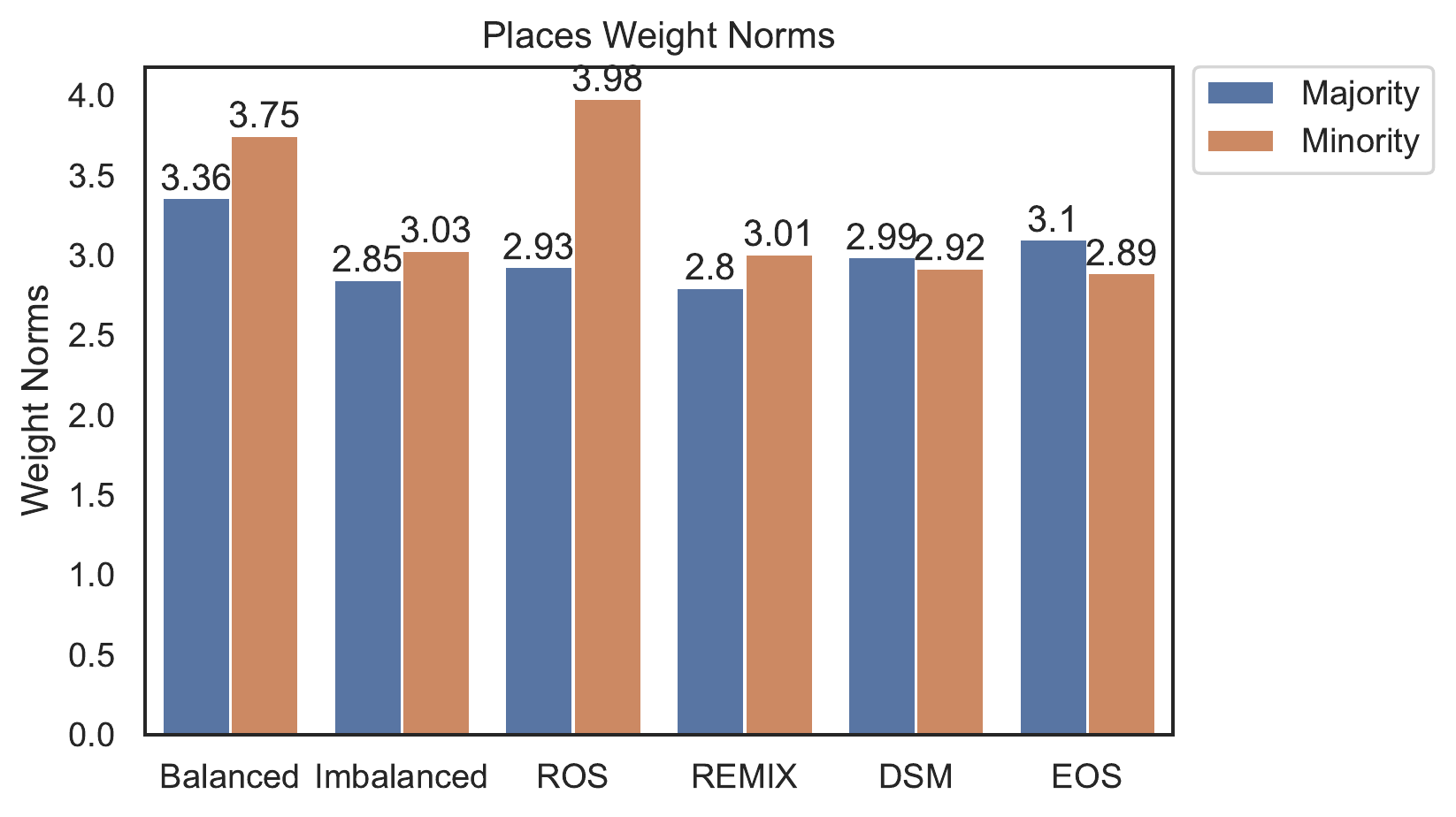}}
\subfloat[INaturalist]
  {\includegraphics[width=0.33\textwidth]{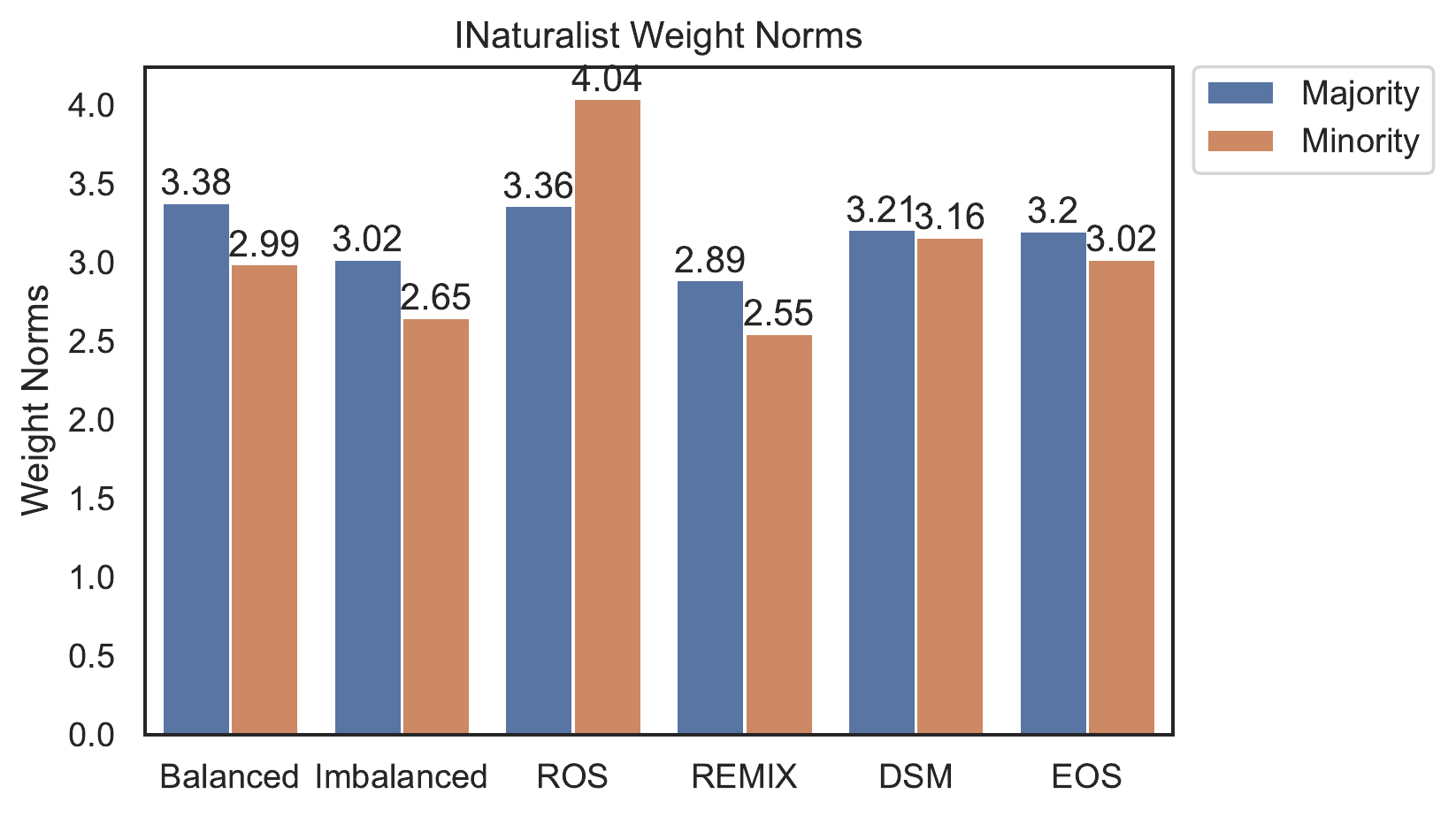}}
  \caption{This figure shows weight norms for 3 image datasets for the majority and minority classes.}
  \label{fig:CIF_wt_norms}
\vspace{-.2cm}
\end{figure*}

Figures~\ref{fig_ch_8_NN_wts_reg_maj} and ~\ref{fig_ch_8_NN_wts_reg_min} show the weight norms for the classification layer of a NN model trained with tabular data for both majority and minority classes. We display the weight norms for the classification layer, instead of the entire model, because we would like to focus on individual class weights. In the case of the NN model, although there are several cases where the weight norms increase (e.g., for SMOTE and ADASYN), for the most part, the general trend for both majority and minority classes is a reduction in weight norms. In contrast to the LG models, this trend appears to clearly indicate weight regularization.

Figures~
\ref{fig:CIF_wt_norms} shows the majority and minority class weight norms for CNN models trained with CIFAR-10, Places and INaturalist data. For CIFAR-10, we treat the class with the largest number of instances as the majority and the one with the least as the minority, since we use exponential imbalance with this dataset. In the case of Places and INaturalist, which use step imbalance, we aggregate the majority classes and select a single minority class.  We also show DA methods that implement augmentation in both real space (ROS and REMIX) and in latent space (DSM and EOS).

We also show the same information for a balanced version of the datasets, so that we can identify whether imbalance is causing a difference in weight norms between the classes, or if other factors, such as data complexity, could be in play.
Although each dataset has a different mix of data complexity (e.g., in some cases, the balanced and imbalanced versions show larger weight norms for the minority class), there are some trends in the DA methods across datasets. 

In the case of the DA methods, there does not appear to be a clear trend with regards to weight regularization. In some cases, the methods increase and other cases they decrease the majority and minority weight norms with respect to their counterparts in the imbalanced base dataset. This is the case whether DA occurs in real space (ROS and REMIX) or in latent space (DSM and EOS). 

\begin{table}[h!] 
\vspace{-.2cm}
\footnotesize
\caption{\textbf{LG Model Weight  Differences}}
\label{tab:ch8_LG_wt_diff}
\centering
\begin{tabular}{ p{1.5cm}p{1.1cm}p{1.1cm}p{1cm}
p{1.1cm}}
\toprule

\textbf{Dataset} & \textbf{ROS} &
\textbf{SMOTE} &
\textbf{ADASYN} & \textbf{REMIX}\\

\midrule
Ozone & 4.172 & 4.795 & 4.938 & 3.817\\
Scene & 3.636 & 3.415 & 3.557 & 4.733\\
Coil & 1.828 & 6.144 & 7.048 & 7.713\\
Thyroid & 1.502 & 2.700 & 2.808 & 2.314\\
US Crime & 8.992 & 8.277 & 9.755 & 8.102\\
\bottomrule

\end{tabular}
\vspace{-.2cm}
\end{table}

To gain further insight into the impact of DA on model weights, we compare the change in model weights between models trained with and without DA. 
 Table~\ref{tab:ch8_LG_wt_diff} shows the difference in weight magnitudes for LG models trained with DA as compared to the same model trained without DA. 

The differences are expressed as the mean of the: weight differences in models trained with and without DA, divided by the weights in the model trained without DA. Thus, it measures the percentage change in each weight between a base model and a model trained with augmented data; and then calculates the mean percentage change of all weights. For all models trained on 5 different datasets, each with 5-way cross validation, the mean change in the weights is greater than 150\% (1.502 for ROS in the Thyroid dataset) and in some cases, over 900\% (9.755 for ADASYN in the US Crime dataset). 

Table~\ref{tab:ch8_NN_wt_diff} presents similar information for models trained with a neural network and tabular data. In this case, the weight changes relate to the classification layer and do not include weight changes in the encoding layers.

\begin{table}[h!]
\vspace{-.2cm}
\footnotesize
\caption{\textbf{NN Model Weight  Differences}}
\label{tab:ch8_NN_wt_diff}
\centering
\begin{tabular}{ p{1.5cm}p{1.1cm}p{1.1cm}p{1cm}
p{1.2cm}}
\toprule

\textbf{Dataset} & \textbf{ROS} &
\textbf{SMOTE} &
\textbf{ADASYN} & \textbf{REMIX}\\

\midrule
Ozone & 3.004 & 2.912 & 2.528 & 2.757\\
Scene & 2.657 & 3.187 & 2.853 & 2.870\\
Coil & 2.738 & 3.037 & 3.023 & 2.343\\
Thyroid & 2.143 & 2.419 & 2.473 & 2.246\\
US Crime & 4.602 & 5.093 & 5.066 & 3.858\\
\bottomrule

\end{tabular}
\vspace{-.2cm}
\end{table}

In all cases, the weights exhibit large changes due to simple numerical equalization of class instances by copying minority class instances (ROS) and also as a result of DA with feature manipulation as a result of interpolation  (SMOTE, ADASYN and REMIX). For all models, the percentage change in the weights ranges from a low of approx. 200\% (2.143) to a high of approx. 500\% (5.093).  

Table~\ref{tab:ch8_img_wt_diff} continues the analysis, but for CNNs trained with image data. We focus solely on the classification layers and exclude the encoding layer weights. Once again, DA induces large changes in weights compared to the base models. Even simple random copying of minority instances induces significant changes in model weights compared to models trained with imbalanced class data (from a low of 200\% - 2.131 - to a high of 2000\% - 20.77).

\begin{table}[h!]
\vspace{-.2cm}
\footnotesize
\caption{\textbf{CNN Weight Differences}}
\label{tab:ch8_img_wt_diff}
\centering
\begin{tabular}{ p{1.7cm}p{ 1.1cm}p{1.1cm}p{1cm}
p{1.2cm}}
\toprule

\textbf{Dataset} & \textbf{ROS} & \textbf{REMIX} &
\textbf{DSM} &
\textbf{EOS}  \\

\midrule
CIFAR-10 & 3.073 & 2.414 & 2.131 & 2.676\\
Places & 5.590 & 2.865 & 3.381 & 3.090\\
Inaturalist & 20.77 & 16.38 & 4.474 & 5.395\\
\bottomrule

\end{tabular}
\end{table}

When these results are taken together, we hypothesize that large changes in weights for LG, NN and CNN models may be evidence of memorization of data points, which is complimentary to our findings with support vectors in SVMs. Random copying of data instances in SVMs increases the number of support vector retained in memory by a SVM model; with the number of SVs greater in cases where DA is implemented with feature manipulation. As synthetic data samples are added, the model requires a larger number of SVs to discern the classes. In the case of LG models, DA results in larger weight norms, which is the opposite of regularization. LG models trained with DA are able to generalize better to the minority class, with some loss of generalization for the majority class. Thus, in LG models, improved generalization with respect to minority classes occurs as a result of larger sized weights.

Both NN and CNN models show reductions in weight magnitudes, which is indicative of regularization; and yet they also experience large changes in weights due to changes in DA. We hypothesize that the large change in weight values is indicative of neuron specialization. As the models are presented with more, synthetic data, with varying feature amplitudes, the classification weights are forced to change their mapping to labels. Weights that reside in last layers of deep models (i.e., the weights in the classification layer that we measure) are forced to change in reaction to data (signal) changes. 

In the next section, we will take a closer look at how DA, when training with imbalanced data, causes front-end changes in feature selection for ML models.

\subsection{RQ4: DA \& front-end feature selection}

\begin{figure}[h!]
   \vspace{-.5cm}
  \centering
  \subfloat[Base]{\includegraphics[width=0.24\textwidth]{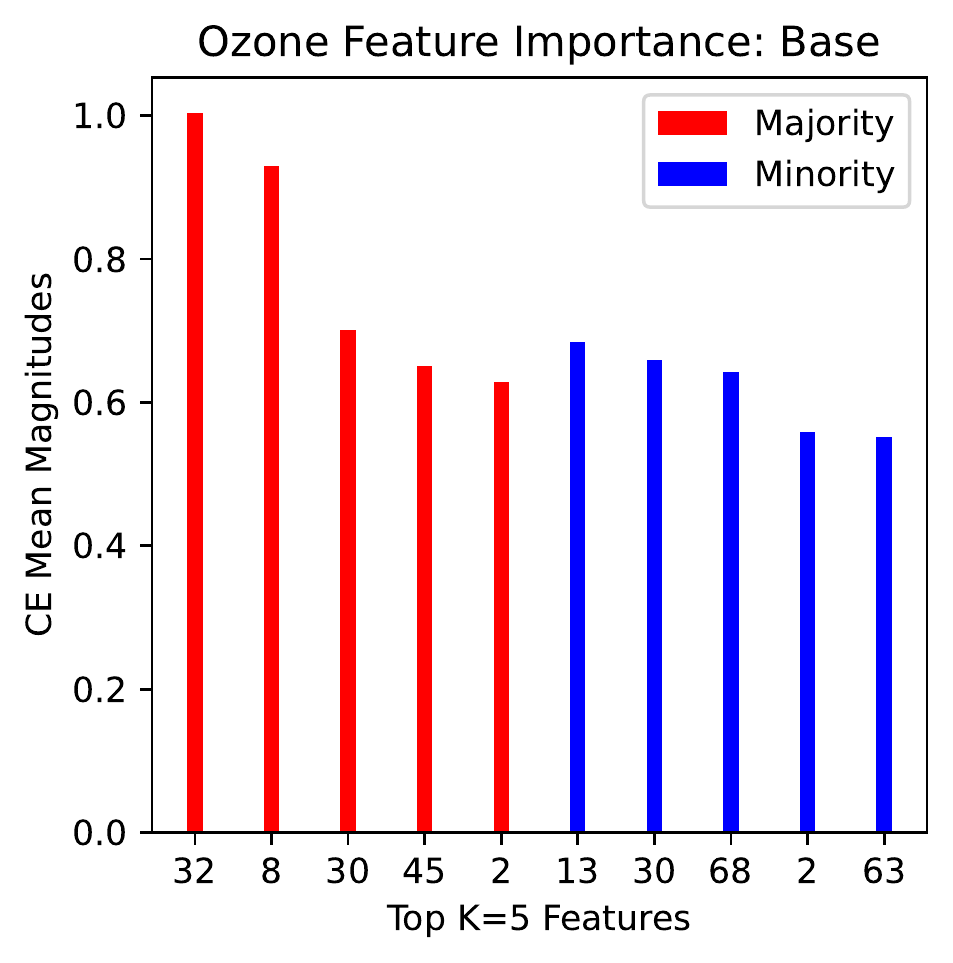}}
  \hfill
  \subfloat[Cost Sensitive]{\includegraphics[width=0.24\textwidth]{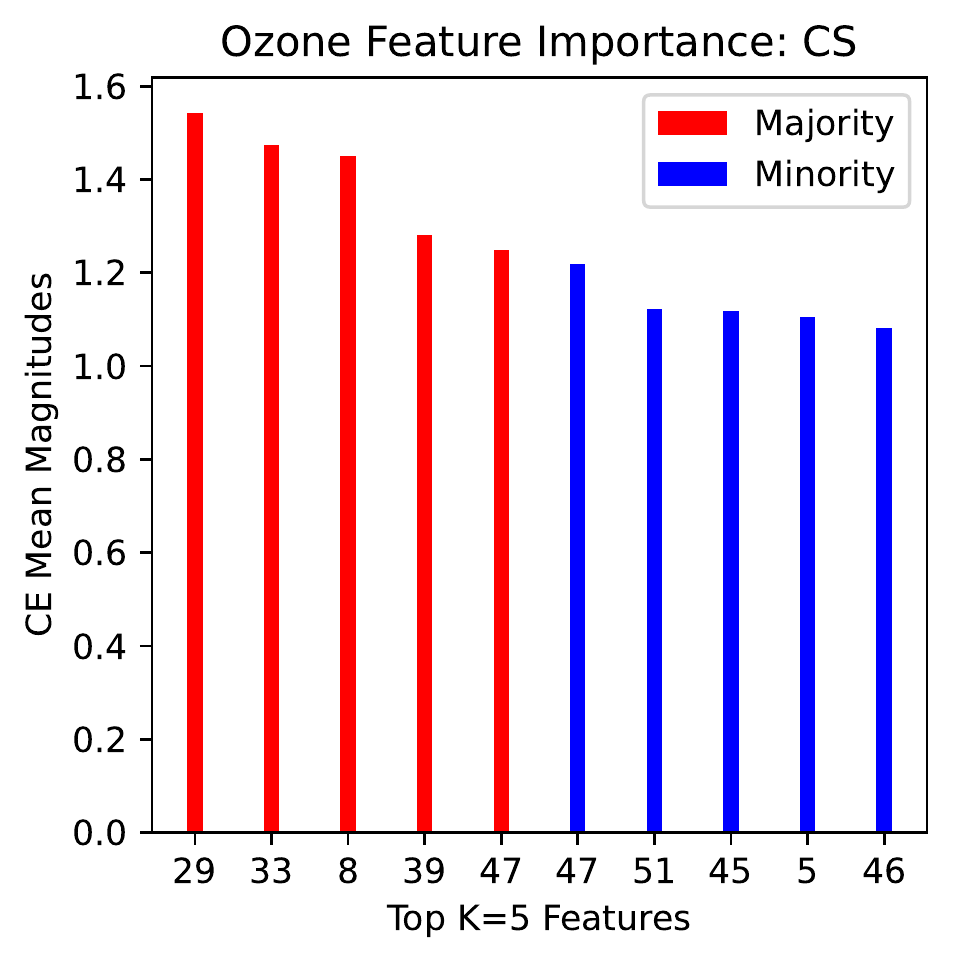}}
  \hfill
  \vspace{-.4cm}
 \subfloat[ROS]{\includegraphics[width=0.24\textwidth]{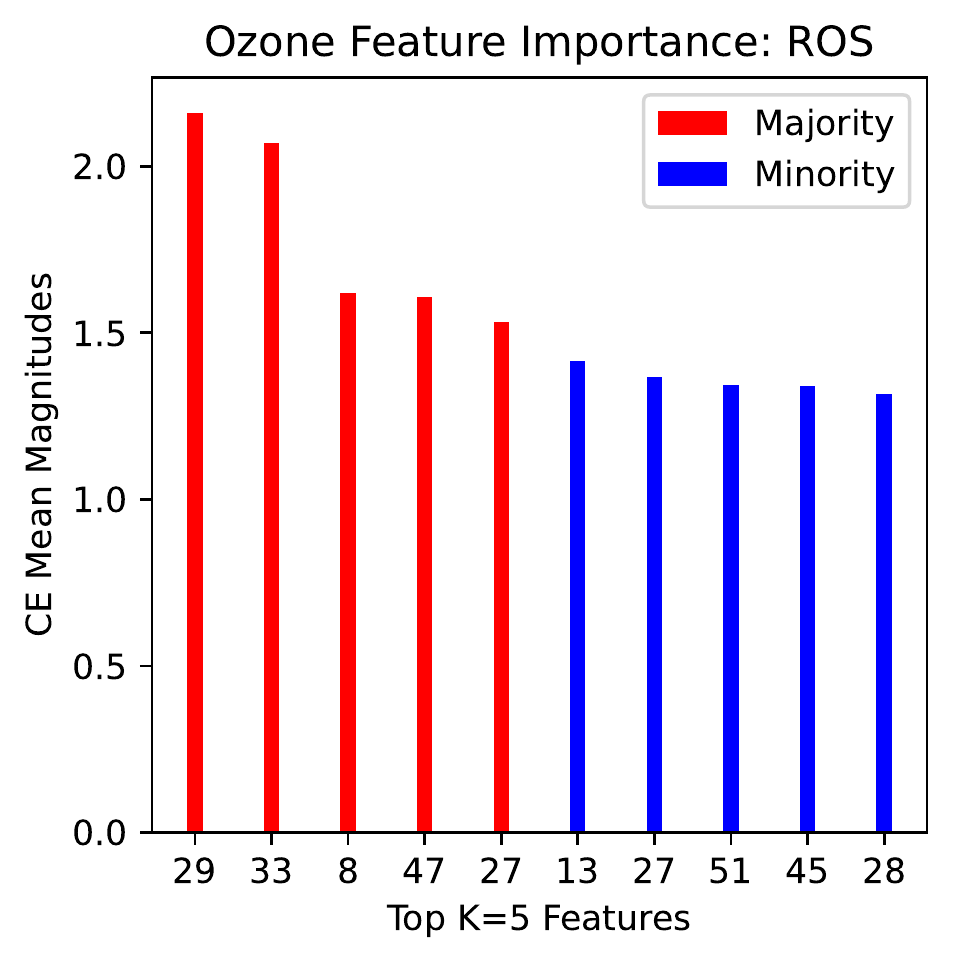}}
  \hfill
  \subfloat[SMOTE]{\includegraphics[width=0.24\textwidth]{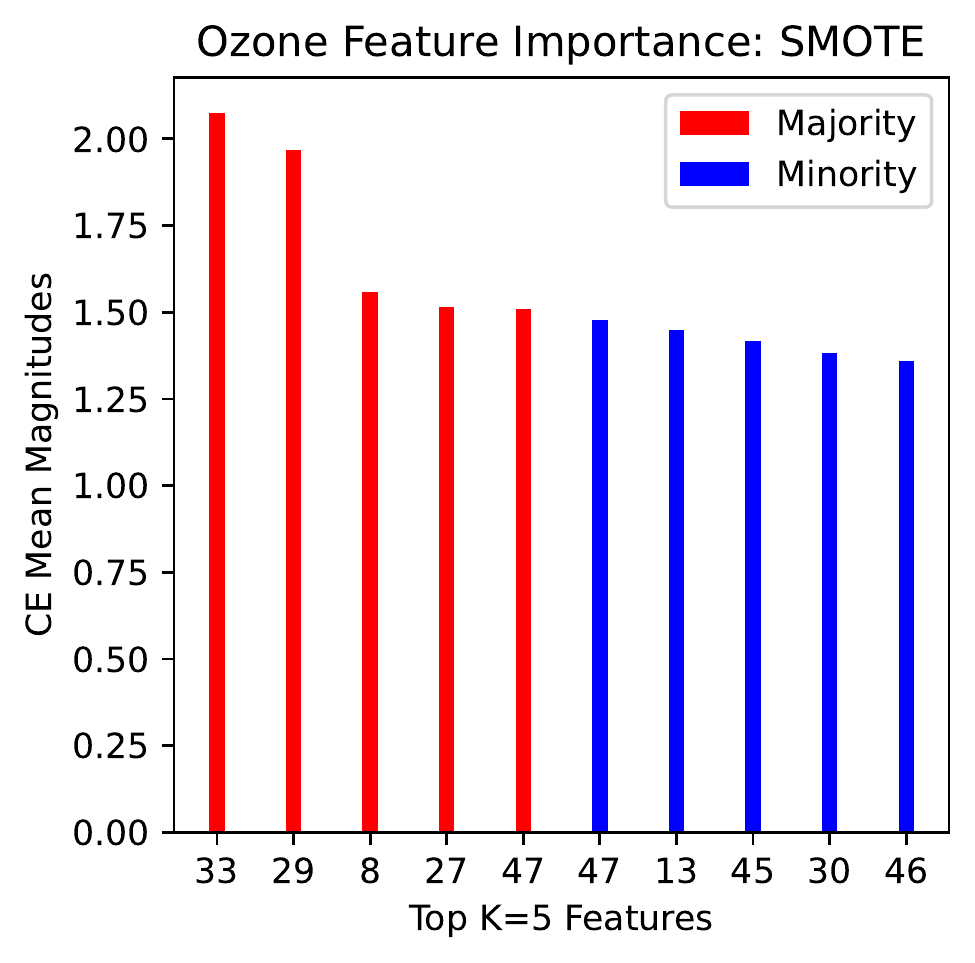}}
  \hfill
  \vspace{-.3cm}
   \subfloat[ADASYN]{\includegraphics[width=0.24\textwidth]{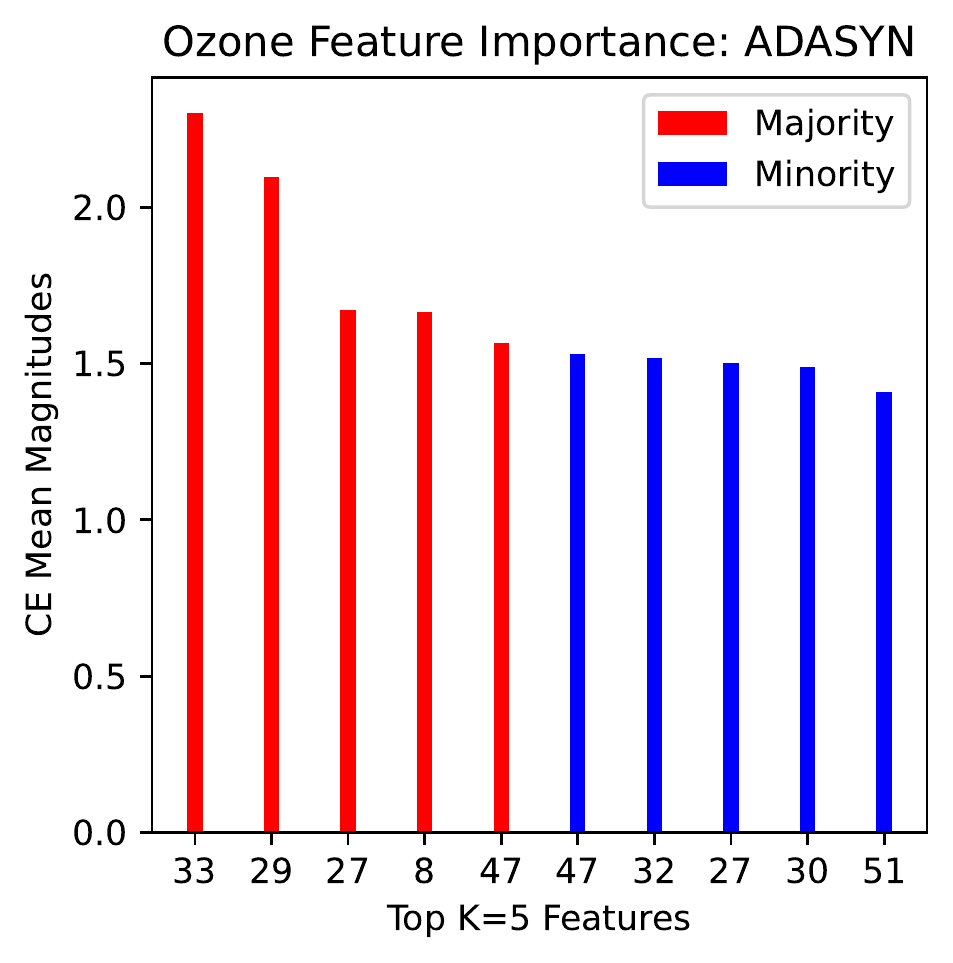}}
  \hfill
  \subfloat[REMIX]{\includegraphics[width=0.24\textwidth]{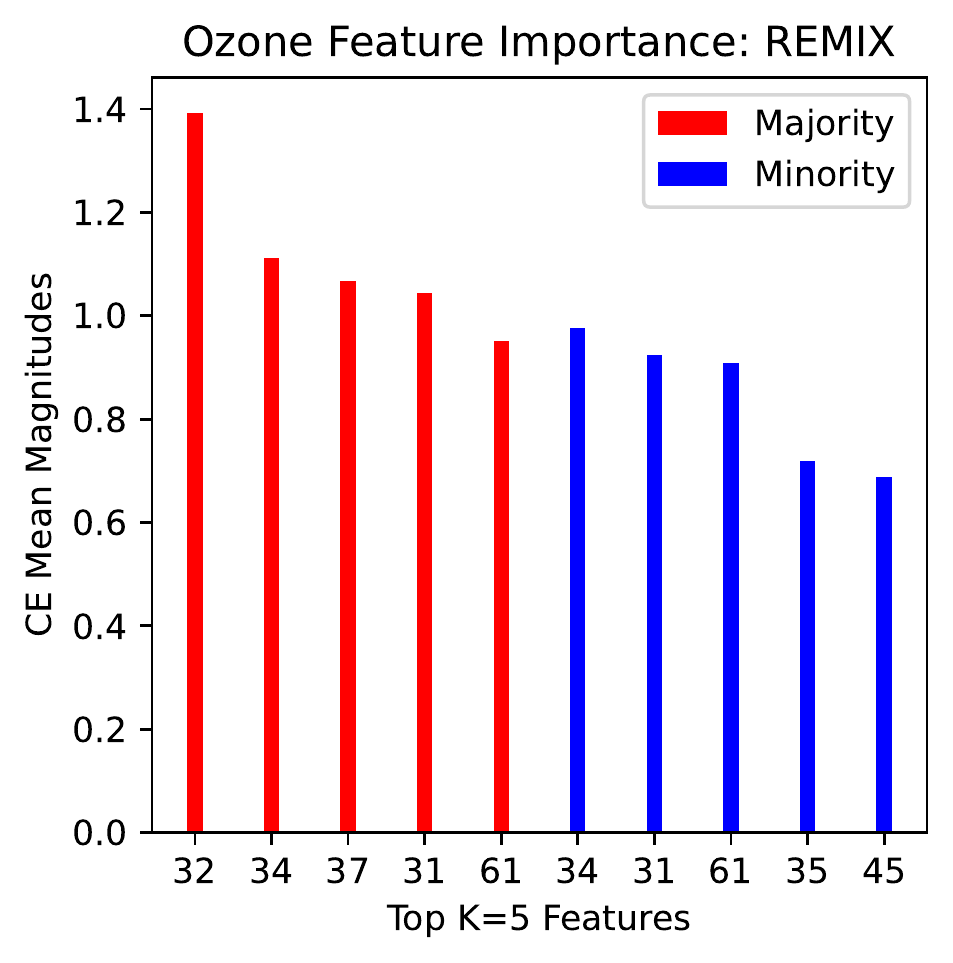}}
  \vspace{-.2cm}
  \caption{The diagram illustrates how DA changes feature importance for LG models trained with and without DA.}
  \label{fig_ch_8_LG_CE_topk_viz}
  \vspace{-.2cm}
\end{figure}

Figure~\ref{fig_ch_8_LG_CE_topk_viz} illustrates how imbalanced learning algorithms change feature importance for LG models trained with tabular data. 

For LG models, we compute feature importance based on the CE mean magnitude of each feature. Because a LG classifier has both positive and negative signed CE when performing binary classification (e.g., positive for the majority class and negative for the minority), the absolute value of each CE feature mean is shown. We visualize the top-5 features, for each class, which have the largest CE mean magnitudes for the Ozone dataset, as well as their individual magnitudes. In general, the majority features (red) have larger magnitudes than the minority (blue). In some cases, the same CE (displayed as an index number of the CE vector on the x-axis) is important for \textit{both} the majority and minority classes, which indicates feature overlap.  For example, in the base model trained with imbalanced data, features \#30 and \#2 are important for both the majority and minority classes in Figure~\ref{fig_ch_8_LG_CE_topk_viz}. 

Most importantly, notice that there is very little overlap in the \textit{identity} of the CE features for each class between the base model, trained with imbalanced data, compared to the other CS and DA algorithms. The identity of the features is listed on the x-axis of each sub-figure.

\begin{table}[h!] 

\centering
\footnotesize
\caption{LG: percentage overlap of top-10 features in models trained with \& without DA}
\label{tab:tabular_overlap}
\begin{tabular}{ p{1.6cm}p{.6cm}p{.8cm}
p{1cm}p{1.2cm}p{1.2cm}
}
\toprule

Dataset & CS & ROS 
& SMOTE & ADASYN & REMIX \\

\midrule
Ozone &  .48 & .43  & .49 & .47 & .39    \\
Scene &  .69 & .68  & .66 & .65 &  .27 \\
Coil & .66 & .63  &  .59 & .58 &  .58    \\
Thyroid & .79 & .73  &  .77 & .78 & .65  \\
US Crime &  .67 & .64  &  .66 & .66 & .52
\\
\midrule
\textbf{Average} &  \textbf{.66} & \textbf{.62}  &  \textbf{.63} & \textbf{.63} & \textbf{.48}
\\
\bottomrule

\end{tabular}
\vspace{-.1cm}
\end{table}

Table~\ref{tab:tabular_overlap} shows the percentage overlap of the 10 features with the largest magnitudes for each class between the base LG model trained with imbalanced data compared to the LG models trained with CS and DA methods. 

The percentage of feature overlap of the top-10 features between the base model and models trained with DA ranges from 27\% to 79\%. The average top-10 feature overlap for models trained with and without DA methods is 60.5\%. This indicates that there is substantial variance (approx. 40\%) in the top-10 features used by models trained with imbalanced data versus augmented data.

\begin{figure}[h!]
   \vspace{-.4cm}
  \centering
  \subfloat[Base]{\includegraphics[width=0.24\textwidth]{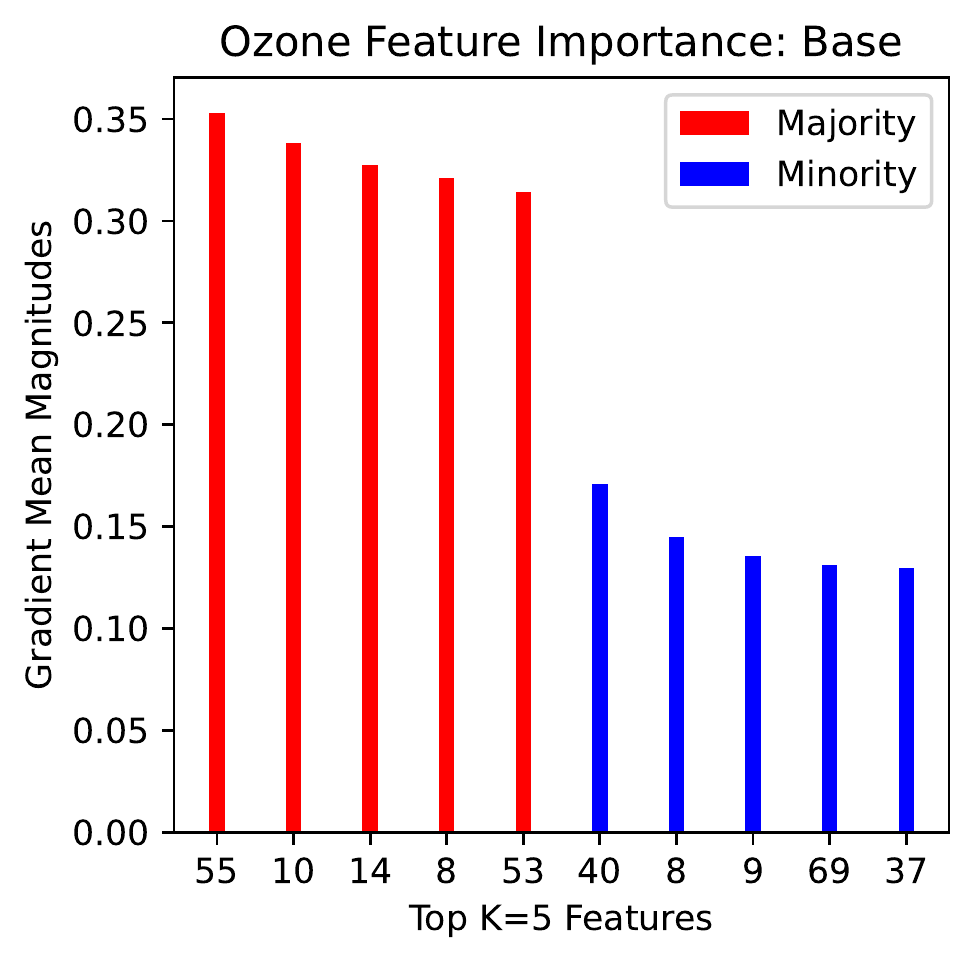}}
  \hfill
  \subfloat[Cost Sensitive]{\includegraphics[width=0.24\textwidth]{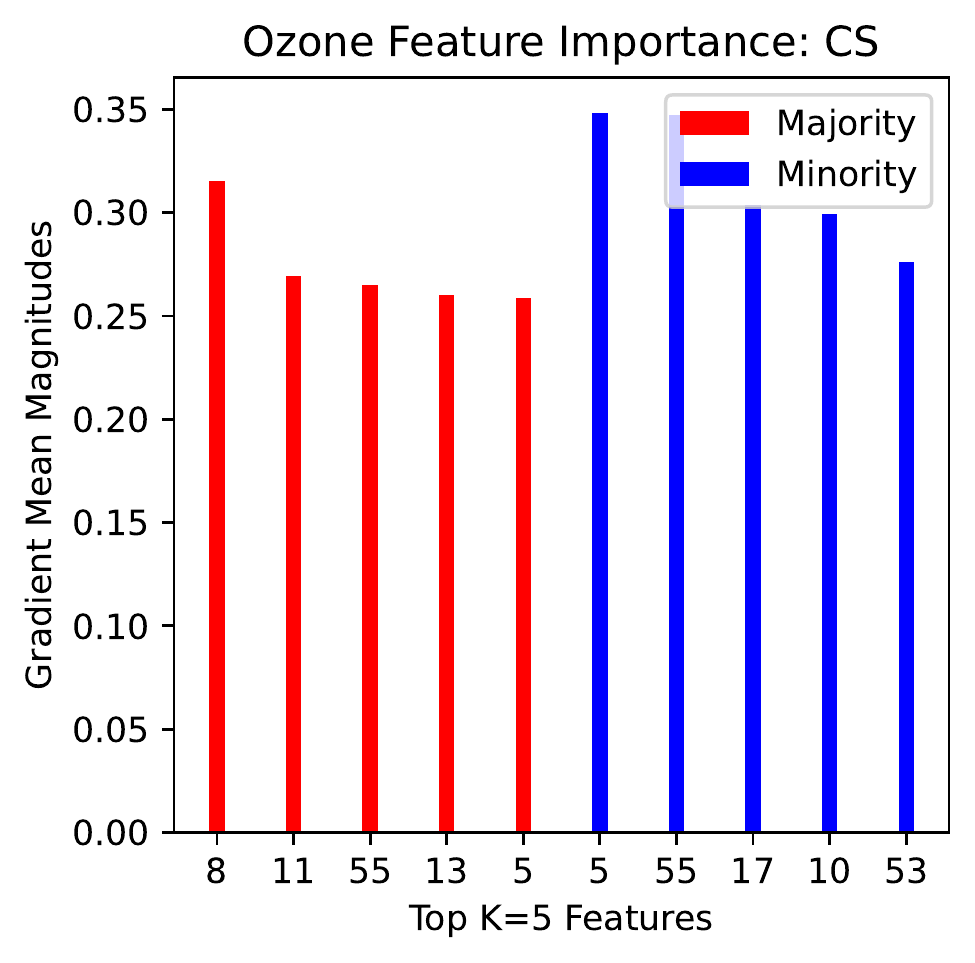}}
  \hfill
  \vspace{-.2cm}
 \subfloat[ROS]{\includegraphics[width=0.24\textwidth]{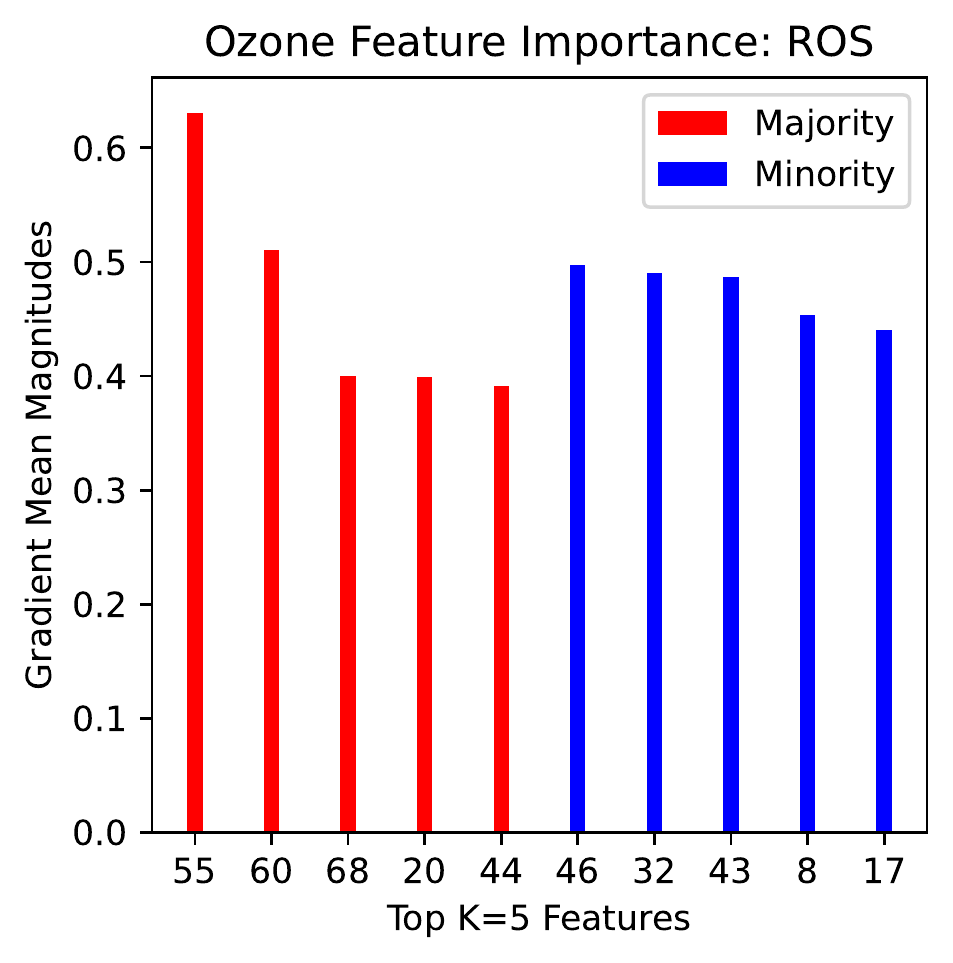}}
  \hfill
  \subfloat[SMOTE]{\includegraphics[width=0.24\textwidth]{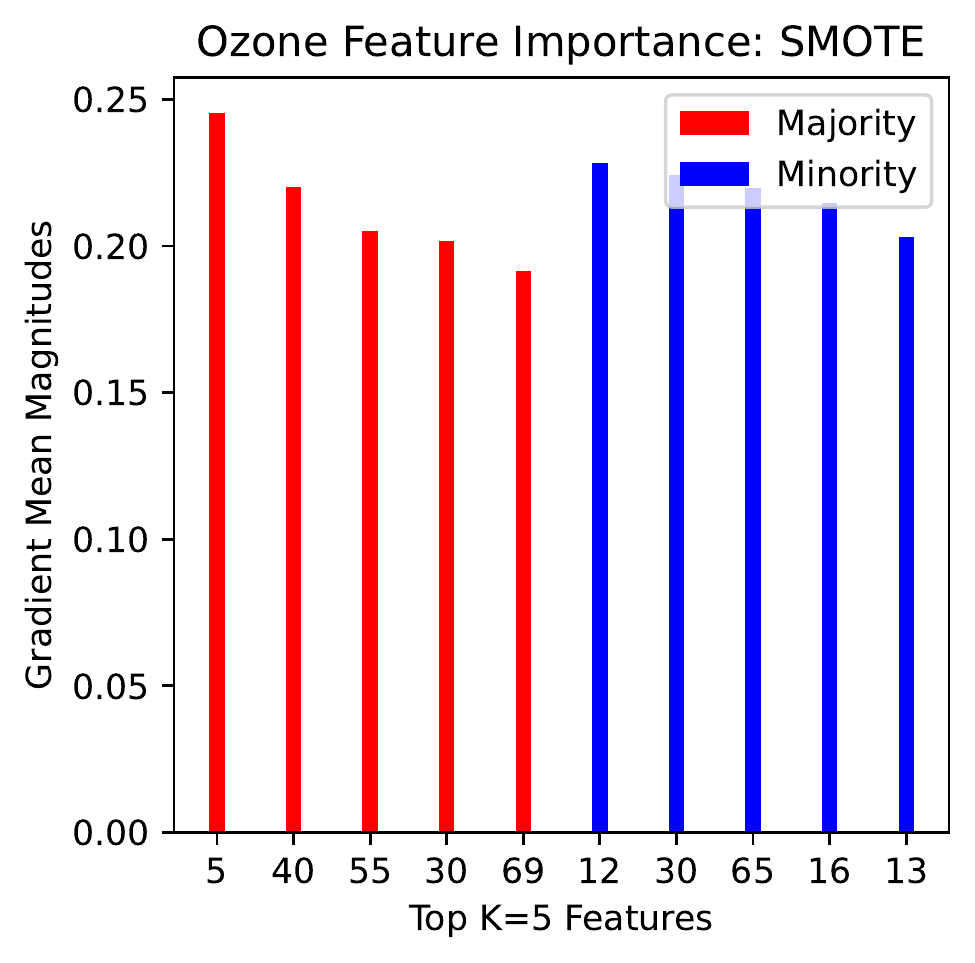}}
  \hfill
  \vspace{-.2cm}
   \subfloat[ADASYN]{\includegraphics[width=0.24\textwidth]{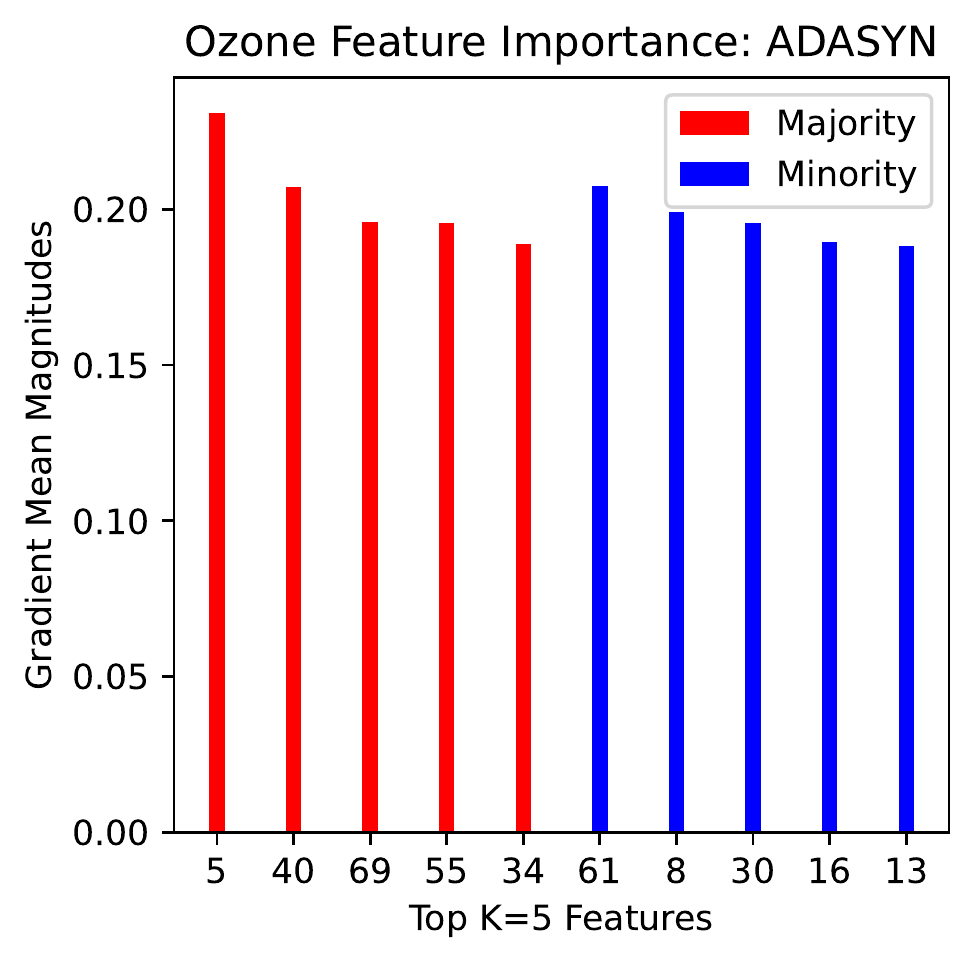}}
  \hfill
  \subfloat[REMIX]{\includegraphics[width=0.24\textwidth]{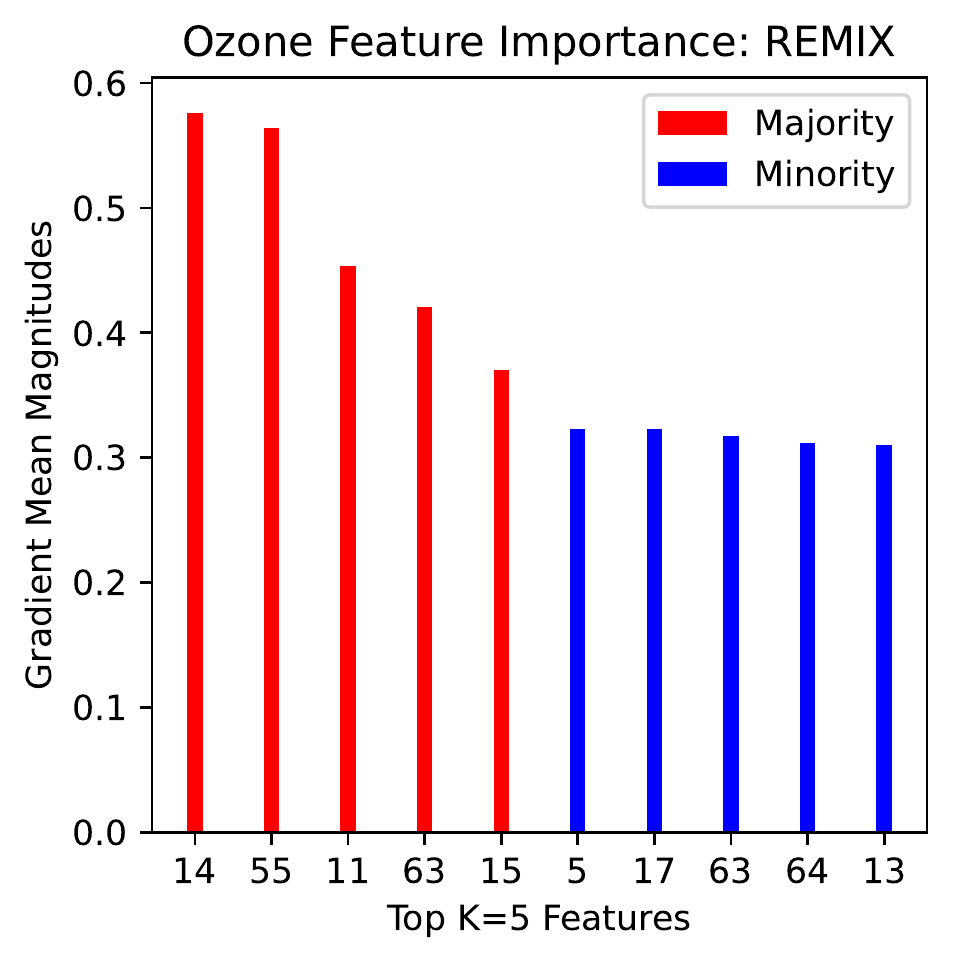}}
  \vspace{-.2cm}
  \caption{The diagram visualizes how DA changes feature importance in tabular data for a neural network model.}
  \label{fig_ch_8_NN_CE_topk_viz}
  \vspace{-.2cm}
\end{figure}

Figure~\ref{fig_ch_8_NN_CE_topk_viz} visualizes how DA changes feature importance in tabular data for a neural network model for the Ozone dataset. Feature importance is computed based on the input gradient mean magnitudes of each feature, which are back-propagated through a trained model. The absolute value of each gradient feature mean is shown. The figure indicates that there is little in the top-5 features for each class in the base model trained with imbalanced data versus models trained with DA. 

\vspace{-.2cm}
\begin{table}[h!] 

\centering
\footnotesize
\caption{NN: percentage overlap of top-10 features in models trained with \& without DA}
\label{tab:NN_tabular_overlap}
\begin{tabular}{ p{1.6cm}p{.6cm}p{.8cm}
p{1cm}p{1.2cm}p{1.2cm}
}
\toprule
Dataset & CS & ROS 
& SMOTE & ADASYN & REMIX \\
\midrule

Ozone &  .56 & .44  & .47 & .46 & .37   \\
Scene &  .22 & .27  & .37 & .32 &  .11 \\
Coil & .50 & .57  &  .56 & .53 &  .52  \\
Thyroid & .68 & .64  &  .64 & .66 & .49 \\
US Crime &  .45 & .48  & .43 & .39 & .32\\
\midrule
\textbf{Average} &  \textbf{.48} & \textbf{.48}  & \textbf{.49} & \textbf{.47} & \textbf{.36}\\
\bottomrule

\end{tabular}
\end{table}

Table~\ref{tab:NN_tabular_overlap} shows the percentage overlap of the top-10 features with the largest magnitudes for each class between the base neural network model trained with imbalanced data compared to the NN models trained with CS and DA methods.

The percentage of feature overlap of the top-10 features between the base model and models trained with DA ranges from 11\% to a high of 68\%. The average top-10 feature overlap for models trained with and without DA is 45.8\%. This indicates that there is substantial variance (approx. 55\%) in the top-10 features used by models trained with imbalanced data and augmented data. 

Finally, Table~\ref{tab:ch8_img_olap} shows the percentage of top-K feature overlap of models trained with and without DA for three image datasets. Feature importance is based on the top 10\% of salient pixels identified through back-propagation. The models trained with a cost-sensitive loss function and real space DA exhibit less feature overlap (approx. 20\% to 27\%) than the models retrained with latent space DA (approx. 75\% to 81\% overlap). The greater salient feature overlap between the base model and the models that are retrained with latent space DA (DSM and EOS) is likely because these models \textit{share the same encoding layers} with the base model. Nevertheless, for all models and datasets, DA triggers a change in feature selection.

\begin{table}[h!]
\footnotesize
\caption{CNN: percentage overlap of top-10 features in models trained with \& without DA}
\label{tab:ch8_img_olap}
\centering
\begin{tabular}{ p{1.3cm}p{1cm}p{1cm}p{1cm}
p{1cm}p{1cm}}
\toprule

\textbf{Dataset} & \textbf{LDAM} & \textbf{ROS} & \textbf{REMIX} &
\textbf{DSM} &
\textbf{EOS}  \\

\midrule
CIFAR-10 & .2519 & .2645 & .2441 & .7891 & .7514\\
Places & .2028 & .2143 & .2342 & .8000 & .7909\\
Inaturalist & .2308 & .2729 & .2721 & .8175 & .8160\\
\midrule
\textbf{Average} & \textbf{.2285} & \textbf{.2506} & \textbf{.2501} & \textbf{.8022} & \textbf{.7861}\\
\bottomrule

\end{tabular}
\vspace{-.1cm}
\end{table}

Collectively, models trained with DA on imbalanced datasets display meaningful changes in model weights and feature selection. The changes occur across model types (SVM, LG, dense NN, CNN) and with both tabular and image data. In some cases, the changes in weights and feature selection is large, for both minority and majority classes, even though DA is only with respect to the minority class. 

These changes occur even when DA for tabular data only involves simple copying of training set minority instances to numerically balance the classes. This triggers a rebalancing of model weights, which causes a change in feature selection (and BAC and F1) for both the majority and minority classes. In the case of DA that incorporates feature manipulation (e.g., through feature interpolation), there is also a change  in model weights and feature selection. 

Both changes, due to simple numerical equalization and noise introduction, cause model weights to pay attention to different features. 

\section{Discussion}
We demonstrate that DA can trigger substantial changes in three parametric ML models (LG, SVM, and CNN) when learning with imbalanced data. In some cases, these changes occur even though DA simply copies training instances to allow for numerical balancing of the classes. 

In the case of a SVM model, DA causes a substantial increase in the number of support vectors required for classification. We hypothesize that more support vectors are required because DA has introduced variance into the training distribution. The variance can be introduced directly, through noise injection by interpolating between training instances. Alternatively, it can be introduced indirectly by causing the model to pay more attention to variation in the minority class by increasing the cost of incorrect prediction.

These results are consistent with Bengio et al.'s observation that \textit{kernel} algorithms, which underpin SVMs, require a \textit{number} of training examples proportional to the number of \textit{variations} \cite{bengio2005curse}. Although they focused on variance in training examples, we believe that this observation also applies to the support vectors learned from the training data. In other words, greater variance in the training data (e.g., through DA) requires a greater number (proportion) of support vectors during inference, when learning with imbalanced data.

We also show that DA can cause large percentage changes in model \textit{weights} in LG and CNN models, when learning with imbalanced data. We speculate that these changes induce modifications in the \textit{paths} by which signals are propagated through single and multi-layer LG and CNN models to the final classification layer. In other words, changing model weights causes changes in the magnitude of signals that propagate through a model. Here, the magnitude of a signal equals the product of the weight and the input to the layer, after an activation function is applied. 

Several authors have analogized neural networks to \textit{kernel} methods. Belkin et al. posit that to understand deep learning requires understanding kernel learning \cite{belkin2018understand}. Jacot et al. find that during gradient descent the parameters (weights) in a neural network follow the \textit{kernel gradient} of function cost \cite{jacot2018neural}. Domingos proposes that neural network's employ \textit{path kernels}, which measure how similar two data points vary during training \cite{domingos2020every}. Zhang et al. show that a neural network is capable of arbitrarily memorizing random data and associating it with a label during training \cite{zhang2021understanding}.

SVMs, which are a form of a kernel method, store a subset of the training set in the form of support vectors. In contrast, LG and CNN models store their learning from data in weights. We conjecture that large percentage changes in model weights as a result of DA may indicate the weights are specializing with respect to \textit{variances} in the data that are associated with a label. Thus, both LG, CNN, and SVM models respond to variances in the data, introduced through DA, by increasing the number of support vectors, or alternatively, weight specialization. Weight specialization may be a form of memorization; just as adding support vectors may be a form of memorization.

\section{Conclusion}

In this paper, we investigate how DA works when learning with imbalanced data through several research questions. First, our research shows that simple numerical balancing of classes, with ROS, contributes a greater share to BAC and F1 improvement than feature manipulation with interpolation. Second, we find that DA in latent space has a greater effect with image versus tabular data.

In addition, our findings demonstrate large changes in model weights, the number of support vectors, and feature selection (the features a model relies on to make decisions) with DA, when training with imbalanced data. Although increasing the amount of synthetic data may yield modest changes in accuracy, it appears to inflict meaningful transformations in  underlying SVM, LG and CNN models, when these models are measured in terms of support vectors, model weights, and feature selection.

\section{Acknowledgements}
This work has been submitted to the IEEE for possible publication. Copyright may be transferred without notice, after which this version may no longer be accessible.

\section{Code \& data}
Python code and data to support representative experiments can be found at \url{https://github.com/dd1github/How_DA_Works}


\bibliographystyle{IEEEtran}
\bibliography{1_refs}

\end{document}